\documentclass[sn-basic]{sn-jnl}

\usepackage{graphicx}%
\usepackage{multirow}%
\usepackage{amsmath,amssymb,amsfonts}%
\usepackage{amsthm}%
\usepackage{mathrsfs}%
\usepackage[title]{appendix}%
\usepackage{xcolor}%
\usepackage{textcomp}%
\usepackage{manyfoot}%
\usepackage{booktabs}%
\usepackage{algorithm}%
\usepackage{listings}%

\usepackage{algorithmic}
\usepackage{array}
\usepackage{textcomp}
\usepackage{stfloats}
\usepackage{url}
\usepackage{verbatim}
\usepackage{subfigure}
\usepackage{booktabs}
\usepackage{multicol} 
\usepackage{arydshln}

\theoremstyle{thmstyleone}%
\theoremstyle{thmstyletwo}%

\theoremstyle{thmstylethree}%

\begin{document}

\title[Article Title]{GASE: Graph Attention Sampling with Edges Fusion for Solving Vehicle Routing Problems}

\author[1]{\fnm{Zhenwei} \sur{Wang}}\email{scxzw2@nottingham.edu.cn}

\author*[1]{\fnm{Ruibin} \sur{Bai}}\email{ruibin.bai@nottingham.edu.cn}

\author[1]{\fnm{Fazlullah} \sur{Khan}}\email{fazl.ullah@nottingham.edu.cn}

\author[2]{\fnm{Ender} \sur{Ozcan}}\email{ender.ozcan@nottingham.ac.uk}

\author[3]{\fnm{Tiehua} \sur{Zhang}}\email{tiehuaz@tongji.edu.cn}

\affil[1]{\orgdiv{School of Computer Science}, \orgname{University of Nottingham Ningbo China}, \orgaddress{\city{Ningbo}, \postcode{315100},\country{China}}}

\affil[2]{\orgdiv{School of Computer Science}, \orgname{University of Nottingham}, \orgaddress{\city{Nottingham}, \postcode{NG8 1BB}, \country{United Kingdom}}}

\affil[3]{\orgdiv{School of Electronics and Information Engineering}, \orgname{Tongji University}, \orgaddress{\city{Shanghai}, \postcode{200092}, \country{China}}}

\abstract{Learning-based methods have become increasingly popular for solving vehicle routing problems due to their near-optimal performance and fast inference speed. Among them, the combination of deep reinforcement learning and graph representation allows for the abstraction of node topology structures and features in an encoder-decoder style. Such an approach makes it possible to solve routing problems end-to-end without needing complicated heuristic operators designed by domain experts. Existing research studies have been focusing on novel encoding and decoding structures via various neural network models to enhance the node embedding representation. Despite the sophisticated approaches applied, there is a noticeable lack of consideration for the graph-theoretic properties inherent to routing problems. Moreover, the potential ramifications of inter-nodal interactions on the decision-making efficacy of the models have not been adequately explored. To bridge this gap, we propose an adaptive Graph Attention Sampling with the Edges Fusion framework (GASE), 
where nodes' embedding is determined through attention calculation from certain highly correlated neighbourhoods and edges, utilizing a filtered adjacency matrix. In detail, the selections of particular neighbours and adjacency edges are led by a multi-head attention mechanism, contributing directly to the message passing and node embedding in graph attention sampling networks. Furthermore, we incorporate an adaptive actor-critic algorithm with policy improvements to expedite the training convergence. We then conduct comprehensive experiments against baseline methods on learning-based VRP tasks from different perspectives. Our proposed model outperforms the existing methods by 2.08\%-6.23\% and shows stronger generalization ability, achieving state-of-the-art performance on randomly generated instances and real-world datasets.}


\maketitle

\section{Introduction}\label{sec1}

The Vehicle Routing Problem (VRP) is a fundamental combinatorial optimization issue in the realm of logistics and transportation management \citep{toth2014vehicle,bai2023analytics}.
VRP belongs to a class of the most challenging combinatorial optimization problems with no proven polynomial-time bounded algorithm yet. 
%
There are two main categories of methods for solving VRP - exact solution methods and approximate optimal solution methods \citep{bai2023analytics}. Exact solution methods seek to build suitable mathematical programming models that are solvable to optimality (at least for problems of smaller sizes) with mainstream methods like branch and bound, branch and pricing \citep{xue2021hybrid,yangBranchpriceandcutAlgorithmVehicle2021}. On the other hand, the approximate solution method uses heuristic or meta-heuristic algorithms with heuristic operators to obtain nearly optimal solutions \citep{chen2020variable}. In recent years, with the continuous development of machine learning, more research moved their attention to the methods of approximately solving VRPs via learning-based end-to-end architecture.

Learning-based methods can be classified into two groups based on different construction solution modes, namely construction heuristics and improvement heuristics \citep{liuHowGoodNeural2023a}. The former employs end-to-end machine learning to construct an approximate optimal solution through autoregression gradually. On the other hand, the latter initially obtains a feasible solution arbitrarily and then leverages machine learning to guide heuristic operators in enhancing the feasible solution to obtain an approximately optimal solution. Different machine learning methods can classify the two solutions mentioned above as either supervised or unsupervised. In supervised methods, the machine imitates the solver's approach during the training process with considerable instances to obtain nearly optimal solutions. An unsupervised learning method can be utilized by employing data-driven strategies, implementing deep reinforcement learning (DRL) \citep{mnihHumanlevelControlDeep2015}, and using a vast amount of learning instances to gradually construct an approximate optimal solution. This can be done through autoregressive encoding and decoding in a trial-and-error manner and increase the probability of effective reward steps through Markov Decision Processes (MDP) \citep{lauriPartiallyObservableMarkov2023}. Alternatively, the feasible solution can be iteratively improved through heuristic operators in reinforcement learning schema to find the approximate optimal solution by heuristic improvement style.Since the inception of pointer networks \citep{vinyalsPointerNetworks2017}, various learning-based methods have emerged to solve VRPs using different network structures, which has pushed machine learning-based approaches to new heights. Some works based on Graph Neural Networks(GNNs) \citep{hamiltonInductiveRepresentationLearning2017a} structure have achieved outstanding results, owing to the powerful node representation learning ability of GNNs. As VRP is a natural graph problem, it can be processed by GNNs not only for nodes but also for abstracting the features of edges between nodes. Even the embedding of the entire graph structure can aid in solving sequences. 

However, supervised learning can be challenging for routing problems like Capacitated VRP(CVRP) in medium and large scale, as obtaining high-quality solutions for labels is not easy. On the other hand, heuristic algorithms require an expert level of domain knowledge to design heuristic operators, and the inference time for new problems can be long. On the other hand, data-driven algorithms don't require the manual design of complex heuristic operators but may result in slightly inferior solutions compared to heuristic algorithms. Therefore, it's essential to develop an unsupervised data-driven intelligent method for solving routing problems with high quality, which is also the motivation behind our work.

This paper describes the use of an unsupervised, end-to-end DRL framework to solve routing problems in an autoregressive manner. The framework uses the actor-critic algorithm to automatically optimize network parameters. To address the exponential growth of search space of the vehicle routing problem, an adaptive sampling graph neural network is constructed using the attention mechanism to combine the representation of nodes and edges. The residual graph network then extracts node features and full graph representation, which are decoded to construct solutions sequentially. To illustrate the performance of the proposed work, we conduct classical experiments on random instances of VRP with nodes of 20, 50 and 100. Furthermore, our evaluation encompasses the time complexity of the algorithm, the speed of model inference, and the model's generalization performance. Our research and experiments revealed that although many studies have achieved impressive outcomes in classic problems, their models' generalization capability requires further improvement. Specifically, we are not aware of any research that examines whether larger-scale end-to-end VRP machine learning models' actual representation is still excellent in small-scale problems, not to mention the reverse. Our experiments confirmed that our proposed novel graph sampling neural network, based on the attention mechanism, performs well in terms of inference speed and generalization performance.

The contributions of this paper can be summarized as the following:

\begin{enumerate}
    \item We present a generic framework to address the vehicle routing problem with capacity constraints using an end-to-end graph learning framework that leverages data-driven patterns. Our framework learns graph representations using efficient encoders and gradually constructs solutions using attention mechanism decoders and masking techniques. We train the encoding and decoding processes with deep reinforcement learning, and the resulting solutions are of high quality without the need for manual heuristic operator design.
      
    \item We propose a novel Residual Graph Attention Sampling neural network that serves as an encoder, which helps improve the node embeddings obtained from sampled essential nodes and edges. This sampling approach is based on a matrix filter that reduces the impact of irrelevant nodes and edges on decision-making time steps according to pair-wised attention scores, leading to a better graph representation, faster convergence of the reinforcement learning process and better generalization performance to various problem sizes.

    \item We employ an adaptive update self-critic policy gradient algorithm to govern the update of model parameters. This involves assessing the significance test values of actors and self-critics in the batch to determine whether the model parameters should be updated. This approach aims to improve the generalizability and credibility of the model results.
    
    \item We conducted extensive verification experiments on classic VRP tasks of varying scales and compared the results with state-of-the-art methods based on various neural network models. Our proposed GASE model outperformed others in terms of solution quality, inference speed, and generalization performance, establishing itself as a leading solution.
    
\end{enumerate}

The rest of the paper is organized as follows. We first summarize the related work in section 2. Section 3 illustrates the preliminaries of the problem. Then we introduce the detailed GASE model and the DRL framework in Section 4. Section 5 presents the experimental results and analysis. Lastly, We wrap up the paper and provide further discussions in Section 6.

\section{Related Work}\label{sec2}

VRP is a combinatorial optimization problem that has gained significant attention in recent years due to its wide range of applications. From simple constraints in the Travelling Salesman Problems(TSP) to more complex ones in the CVRP and the Vehicle Routing Problem with Time Windows (VRPTW), VRPs have become a series of well-known NP-hard problems. In the last few decades, various solution algorithms have been developed to find the optimal solution for the global route on small instances, such as mixed integer programming, 
dynamic programming, 
branch pricing, 
and other methods. \citep{braekers2016vehicle} However, due to the NP-hard nature of VRP, the exact solution algorithm faces the issue of combinatorial explosion as the problem's scale grows. It becomes increasingly difficult to obtain the optimal solution within polynomial time complexity. For this reason, approximate algorithms aim to find near-optimal solutions and have become the subject of intense research by scholars in recent years. One direction for approximate algorithms to solve VRP is to use heuristic or meta-heuristic algorithms, such as evolutionary multitasking algorithm \citep{fengExplicitEvolutionaryMultitasking2021b}, artificial bee colony algorithm \citep{ngMultipleColoniesArtificial2017}, variable neighbourhood search algorithm \citep{kalatzantonakisReinforcementLearningVariableNeighborhood2023a}, etc. Another branch is supervised or DRL methods represented by machine learning. A challenge with supervised methods is that they require accurate labels for training instance data, which are typically obtained from high-quality solutions from optimal or heuristic solvers. Among the representative works, Vinyal et al. \citep{vinyalsPointerNetworks2017} proposed using pointer networks(PtrNet), that is, the sequence model of Recurrent Neural Networks(RNNs) \citep{zarembaRecurrentNeuralNetwork2015}, to train TSP instances end-to-end, thereby obtaining an approximately optimal solution for untrained data or test data. The emergence of pointer networks has pushed the use of machine learning models to solve routing problems to a new climax.

The following paragraphs will review DRL-based routing methods that can be categorized as construction heuristics or improvement heuristics.
\subsection{DRL based construction heuristic}
The construction heuristic model utilizes neural networks to perform end-to-end encoding and decoding. The encoder extracts node features while the decoder combines these extracted features to output the selection of node sequences by the neural network at different time steps. Through reinforcement learning, the neural network structure is trained to increase the probability of outputting high-profit nodes at the current time step. This autoregressive approach constructs the entire process step by step, eliminating the need for manually designing complex heuristic operators. 
\citep{belloNeuralCombinatorialOptimization2017} employed the pointer network structure and DRL approach along with the actor-critic algorithm to train the PtrNet model to solve TSP. Each training instance was treated as a sample, with the negative value of the entire trajectory length serving as the reward. The policy gradient was calculated using Monte Carlo estimation to obtain the output of near-optimal solutions. \citep{nazariReinforcementLearningSolving2018} proposed a network based on Long Short Term Memory(LSTM) and attention mechanism, which performs better on medium-size VRP compared with Google OR-Tools \citep{ortools}. 
\citep{koolAttentionLearnSolve2019} raised a model based on Transformer architecture \citep{vaswaniAttentionAllYou2023}, called the Attention Model(AM), which outperforms PtrNet. In addition, the REINFORCE algorithm and the deterministic greedy algorithm are used as baselines. It can be a near-optimal solution in the tsp with 100 nodes. Using the same hyperparameters, VRP performance is better than baseline and close to optimal algorithms, making it state-of-the-art. 
\citep{kwonPOMOPolicyOptimization2020} designed a Policy Optimization with Multiple Optima (POMO) schema, which is an innovative training method based on end-to-end architecture, using the modified REINFORCE algorithm to force different deployments for all the best solutions. Its low variance accelerates the training and stability of RL, with more capability of resisting local minima.
In addition to the sequence structure, another line chooses to use GNNs to process VRP. For example, \citep{khalilLearningCombinatorialOptimization2017} uses struct2vec network coding and deep Q-network for training. 
\citep{leiSolveRoutingProblems2022} proposed an end-to-end model based on the graph attention network structure, decoding through the transformer architecture, using the PPO algorithm for training, and achieving state-of-the-art performance on small and medium-sized TSP and VRP. 
\citep{joshiLearningTravellingSalesperson2022} presented an end-to-end neural combinatorial optimization pipeline via GNNs to identify the graph embedding and autoregressive decoding process and verify the generalization ability of graph embeddings for TSP. 
The studies mentioned above have achieved promising results in solving TSP and VRP. Also, the time required for inference is minimal owing to the characteristics of end-to-end machine learning. However, these models are heavily reliant on pure data driving, which limits the further quality of the solutions obtained. Additionally, these models tend to be heavily influenced by the data distribution of training instances, leading to suboptimal performance when dealing with data of varying scales and distributions. 
\subsection{DRL based improvements heuristic}
In recent years, the use of the RL framework to handle the stepwise optimization process of a feasible solution by heuristic operators has also been sought after by many studies.  
\citep{maCombinatorialOptimizationGraph2019} trained a graph pointer network (GPN) by hierarchical reinforcement learning to solve TSP with complex constraints. At the same time, the solution obtained by the GPN framework can be further combined with local search such as the 2opt operator to obtain a further optimized solution. 
\citep{zhaoHybridDeepReinforcement2021} utilized a routing simulator, and actor-adaptive critics algorithms to build a DRL framework. The model receives graph information via the simulator and generates routing strategies via the attention mechanism. The solutions produced are then combined with various local search methods for further improvements. 
The aforementioned research has proposed different innovations and highlights in the combination of DRL and heuristic algorithms. Nonetheless, heuristic algorithms still hold significant sway, thereby necessitating the acquisition of expertise in the relevant domain and the development of some manually designed heuristic operators.

\section{Problem Preliminaries}
In this section, we will formulate the typical VRP and introduce the preliminaries of our DRL framework. It should be noted that, given the Traveling Salesman Problem (TSP) represents a more rudimentary variant of the VRP, the present discourse will not delve into an extensive elaboration of the former. For comprehensive expositions of TSP, readers are directed to consult the subsequent scholarly article \citep{vinyalsPointerNetworks2017,koolAttentionLearnSolve2019,leiSolveRoutingProblems2022}, which provides a standardized elucidation of the subject matter.

In VRP, a vehicle departs from a fixed depot and traverses a series of customer nodes, collecting goods from each customer along the way and subsequently returning them to the depot. Given a VRP graph $ G= (V, E) $, where $V =\{v_0,v_1,...,v_n\}$ represents the set of vertex, and $E = \{(v_i,v_j)\mid v_i,v_j\in V,i\neq j\}$ denotes the possible edges between pairs of nodes. To address the connectivity of node pairs, we use an adjacency matrix $A \in\{0,1\}^{|V| \times |V|}$ where $A_{i,j} = 1 $ if $(v_i,v_j)\in E$ and $A_{i,j} = 0 $ otherwise, $|\cdot|$ indicates the node sequence, that is, the number of nodes. For each customer node $i$, $c_i = (x_i,y_i)$ indicates the Euclidean coordinate of node $v_i$ where $q_i$ is the demand. Note that $v_0$ represents the depot with demand 0. The vehicle will visit every customer exactly once while the total demands of nodes must not exceed the vehicle max load $Q$. The detailed information of variables is shown in Table \ref{tabvar}, the object and constraints of VRP are formulated as follows:

\begin{table}[h!]
\caption{Variable and Notation Definition}
\label{tabvar}
\centering
\begin{tabular}{ll}
\toprule
Variable or Notation & Definition\\
\hline
$V =\{v_0,v_1,...,v_n\}$ & set of vertex, $v_0$ for depot\\
$n$ & number of customer nodes\\
$E = \{(v_i,v_j)|v_i,v_j\in V,i\neq j\}$ & set of edge\\
$A \in\{0,1\}^{|V| \times |V|}$ & adjacency matrix\\
$\mathcal{N}\left(v_i\right)$ & neighborhoods of $v_i$\\
$c_i = (x\_value,y\_value)$ & euclidean coordinate of node $i$\\
$Q$ & max load of the vehicle\\
$Q_{c}$ & current load of the vehicle \\
$t$ & current time step\\
$Q_t$ & vehicle remaining capacity at step t\\
$q_{i}, 0\leq q_{i}<Q$ & initial demand of node $i$ \\
$q_{i,t}, 0\leq q_{i,t}<Q$ & demand of node $i$ at step t\\
$d_{ij}$ & distance between node $i$ and $j$\\
$r_{ij}$ & binary decision variable to indicate edges in solution \\
$\mathcal{K}$ & vehicle numbers\\
$\pi$ & solution sequence \\
$\pi_t$ & solution node at step t \\
$S$ & current state of the DRL environment \\
$\theta$ & neural network parameters\\
$p_{\theta}\left(\cdot\right)$ & probability under neural network parameter $\theta$ \\
$L$ & total solution length\\
$x_i$ & feature embedding of node $i$\\
$e_{ij}$ & feature embedding of edge $e_{i,j}$ \\
$W_*^*$ & learnable weight matrices \\
$b_*^*$ & bias\\
$K$ & hyperparameter for sampling in encoder\\
$K\%$ & sampling rate in encoder\\
$h_i^l$ & hidden state of node $i$ at graph residual connection layer $l$ \\
$H^l$ & hidden state of all graph nodes at graph residual connection layer $l$ \\
$H^L$ & output state of all graph nodes after a $L$-layer residual connection \\
$V_{agg}^l$ & node sets to do feature aggregation at residual connection layer $l$\\
$V_{i\_agg}^l$ & neighbor sets of node $i$ after filter at residual connection layer $l$ \\
$\alpha_{ij}^l$ & attention coefficient from node $j$ to node $i$ at residual connection layer\\
$Z(g)$ & whole graph readout\\
$H$ & number of attention head \\
$c_t^m$ & context vector of multi-head attention layer in decoder \\
$u_*^m$ & attention coefficient of multi-head attention layer in decoder \\
$c_t^s$ & context vector of single-head attention layer in decoder \\
$u_*^s$ & attention coefficient of single-head attention layer in decoder \\

\bottomrule
\end{tabular}
\end{table}

\begin{equation}\label{eq1}
    \min \sum_{i \in V} \sum_{j \in V} d_{i j} r_{i j} ,\quad \forall i, j \in V,r_{i j} \in\{0,1\} 
\end{equation}

subject to

\begin{equation}\label{eq2}
    \sum_{i \in V} r_{i j}=1 ,\quad \forall j \in V \backslash\{0\}
\end{equation}

\begin{equation}\label{eq3}
    \sum_{j \in V} r_{i j}=1 ,\quad \forall i \in V \backslash\{0\}
\end{equation}

\begin{equation}\label{eq4}
    \sum_{i \in V} r_{i 0}=\mathcal{K}
\end{equation}

\begin{equation}\label{eq5}
    \sum_{j \in V} r_{0 j}=\mathcal{K}
\end{equation}

\begin{equation}\label{eq6}
    \sum_{i \notin P} \sum_{j \in P} r_{i j} \geq r(P) ,\quad \forall P \subseteq V \backslash\{0\}, v \neq \emptyset
\end{equation}
Note that $r_{ij}$ is a binary decision variable that indicates whether the $e_{\left(i,j\right)}$ is part of the solution and $d_{ij}$ is the cost(distance) of using $e_{\left(i,j\right)}$. $\mathcal{K}$ is the number of vehicles being used(could be one for multiple routes) and $r\left(P\right)$ is the minimum number of vehicles required to serve customer set. Constraints (\ref{eq2} and \ref{eq3}) make each customer is visited exactly once and constraints (\ref{eq4} and \ref{eq5}) ensure the satisfaction of the number of vehicle routes. Finally, constraint (\ref{eq6}) makes sure that the demands from all customers are fully satisfied. Then given a solution $\pi$ under the VRP environment $G$, the object equation for DRL schema is to minimize the total solution length $L(\pi \mid G)$:

\begin{equation}\label{eq7}
    L(\pi \mid G)=\sum_{i=0}^{|\pi|}\left\|c_{\pi(i)}-c_{\pi(i+1)}\right\|_2
\end{equation}

where $\left\|\cdot\right\|$ computes the L2 distance of node pairs, $c_0 = c_{\pi+1}$ and $|\pi|$ represents for the sequential length of solution $\pi$. Note that $|\pi| > n $ as depot $v_0$ may appear multiple times during the solution trajectory for VRP but only at the beginning and end for TSP. The following are the fundamental components of the overall DRL architecture:
\begin{enumerate}
    \item \textit{State}:
    
    The system states are observed by two main characteristics: the location and demand of the customer node, and the location and loading of the vehicle at the current time step. The characteristics can be classified into two categories: dynamic and static. The static features include the location features (Euclidean coordinates $c_i$). On the other hand, the dynamic features change over time. The vehicle's load $Q_c$ increases or clears (when it returns to the depot $v_0$), while the customer's demand characteristics $q_i$ are cleared once the vehicle visits them.

    \item \textit{Action}:

   The system environment analyzes the current status of the time step and suggests the next node to be visited as an action to the vehicle. The unvisited nodes are available as actions for the vehicle, and the system's strategy determines the probability distribution for implementing actions. The current action at time step t is noted as $\pi_t$ while the conditional probability is $p(\pi_t\mid \pi_1...\pi_{t-1}, S)$, where $S$ is the current state.

   \item \textit{Reward}:
   
   To calculate the reward in reinforcement learning, we take the negative value of the path length of the node that is visited after acting. This is because we want to optimize for the shortest access path and hope that the reward will increase. After executing a strategy based on the probability distribution, the overall return is calculated as the negative value of the path length after all node conditions are met. In other words, it is equal to $- L(\pi \mid G)$.
\end{enumerate}

To summarize, the process of producing solution $\pi$ is determined by the system strategy, whose composition is the multiplication of the probabilities of selecting actions at different time steps $t$ for stochastic policy. The training process is expected to adjust the neural network parameter $\theta$ to find a solution under the given CVRP graph $G$. The strategy can be expressed as follows:

\begin{equation}\label{eq8}
    p_\theta(\pi \mid G)=\prod_{t=1}^m p_\theta\left(\pi_t \mid G, \pi_{t^\prime}, \forall t^\prime<t\right)
\end{equation}

\begin{figure*}[h!]
\centering
\includegraphics[width=1.0\textwidth]{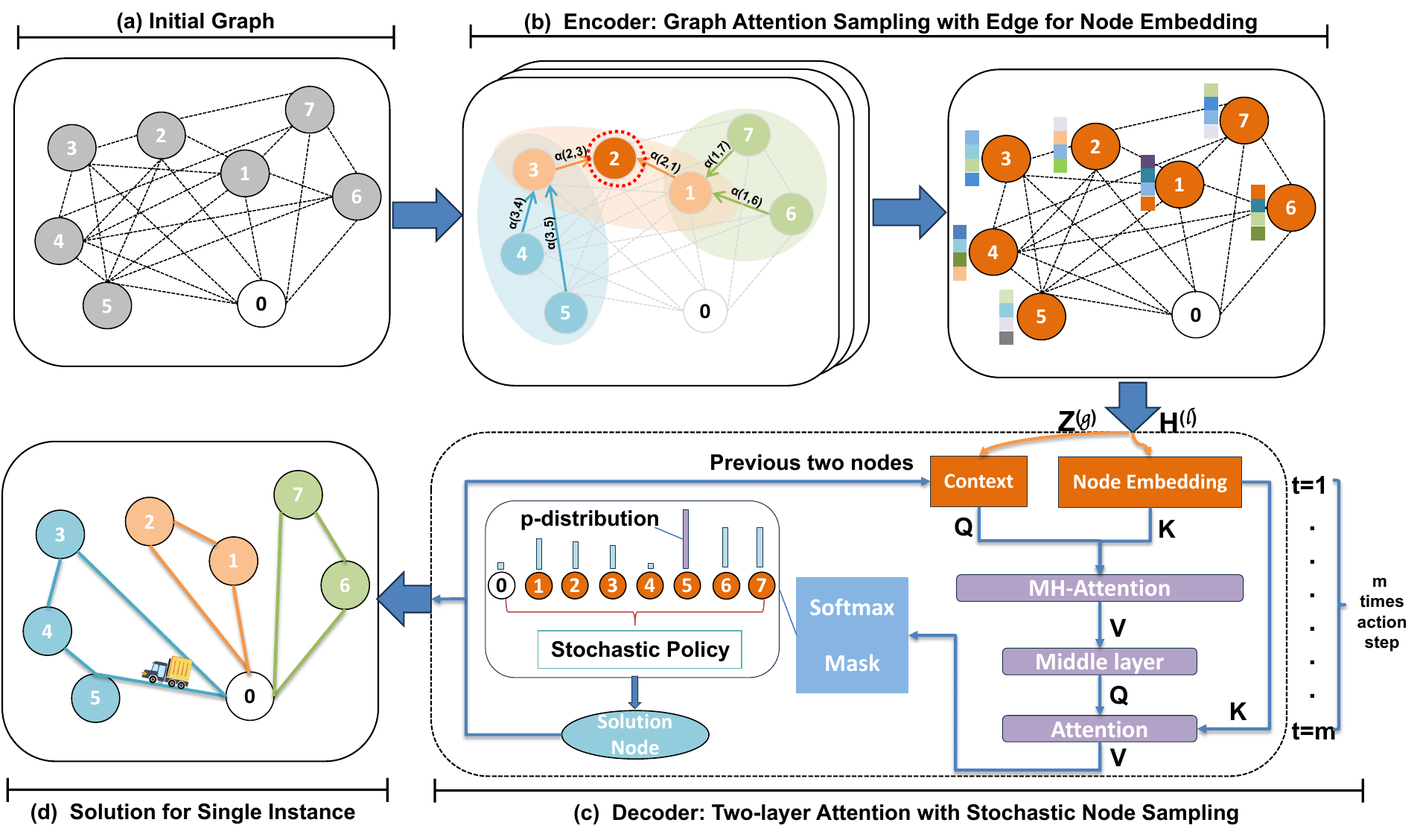}
\caption{An end-to-end GASE schema pipeline}
\label{figure1}
\end{figure*}

\section{GASE Model}
In this section, we will demonstrate the GASE model pipeline, including encoder feature representation with attention sampling strategy, decoder solution generalization, and training algorithms.

\subsection{GASE framework}
Figure~\ref{figure1} illustrates the process of using an end-to-end architecture to generate viable solutions for VRP instances. In Figure \ref{figure1}(a), the initial graph of a VRP instance is presented. To complete the input to Figure \ref{figure1}(b), features of nodes and edges are linearly mapped through a fully connected layer and a batch normalization \citep{ioffe2015batch} layer BN as shown in Eq.(\ref{eq9}) and Eq.(\ref{eq10}). Here, we use separate parameters $W_{dp}$ and $b_{dp}$ to represent the initial embedding of the depot node.

\begin{equation}\label{eq9}
    x_i = \left\{
    \begin{array}{ll}
       BN(W_{dp}c_i+b_{dp})  & i = 0 \\ 
       \\
       BN[W_0(c_i||q_i)+b_0]  &  i = 1...n\\
    \end{array}
    \right.
\end{equation}

\begin{equation}\label{eq10}
    e_{ij} = BN(W^{\prime}_0 d_{ij}+b^{\prime}_0)
\end{equation}
The features of customer node $i (i=1...n)$ include its deterministic coordinates and dynamic demands (customer demands change once visited), while the Euclidean distance between node pairs $i,j$ (i.e. $d_{i,j}$) represents the embedding of edges, $\cdot||\cdot$ is a concatenation operation. When doing message passing in Graph Attention Networks (GAT)\citep{velickovicGraphAttentionNetworks2018a}, we only consider aggregating the features of important neighbours using graph attention sampling with a matrix filer, which is shown in Fig. \ref{figure1}(b). The encoder produces node representation and the entire graph readout after multiple residual graph neural networks, which will be discussed in detail in the following subsection. The decoder uses an attention mechanism to gradually output the node sequence through the embedded representation of the context and graph nodes. The context representation draws on the recurrent neural network. It will be updated with time steps to the node embedding of the entire graph readout and the node representation output in the previous two action steps. The output node is sampled from a probability distribution to avoid getting stuck in local optima during DRL training, nodes that are unsatisfied will be masked to make probability vanish as shown in Fig. \ref{figure1}(c). The decoding process will repeat m times to produce a complete solution for a VRP instance like Fig. \ref{figure1}(d).

\begin{figure*}[ht]
\centering
\includegraphics[width=1.0\textwidth]{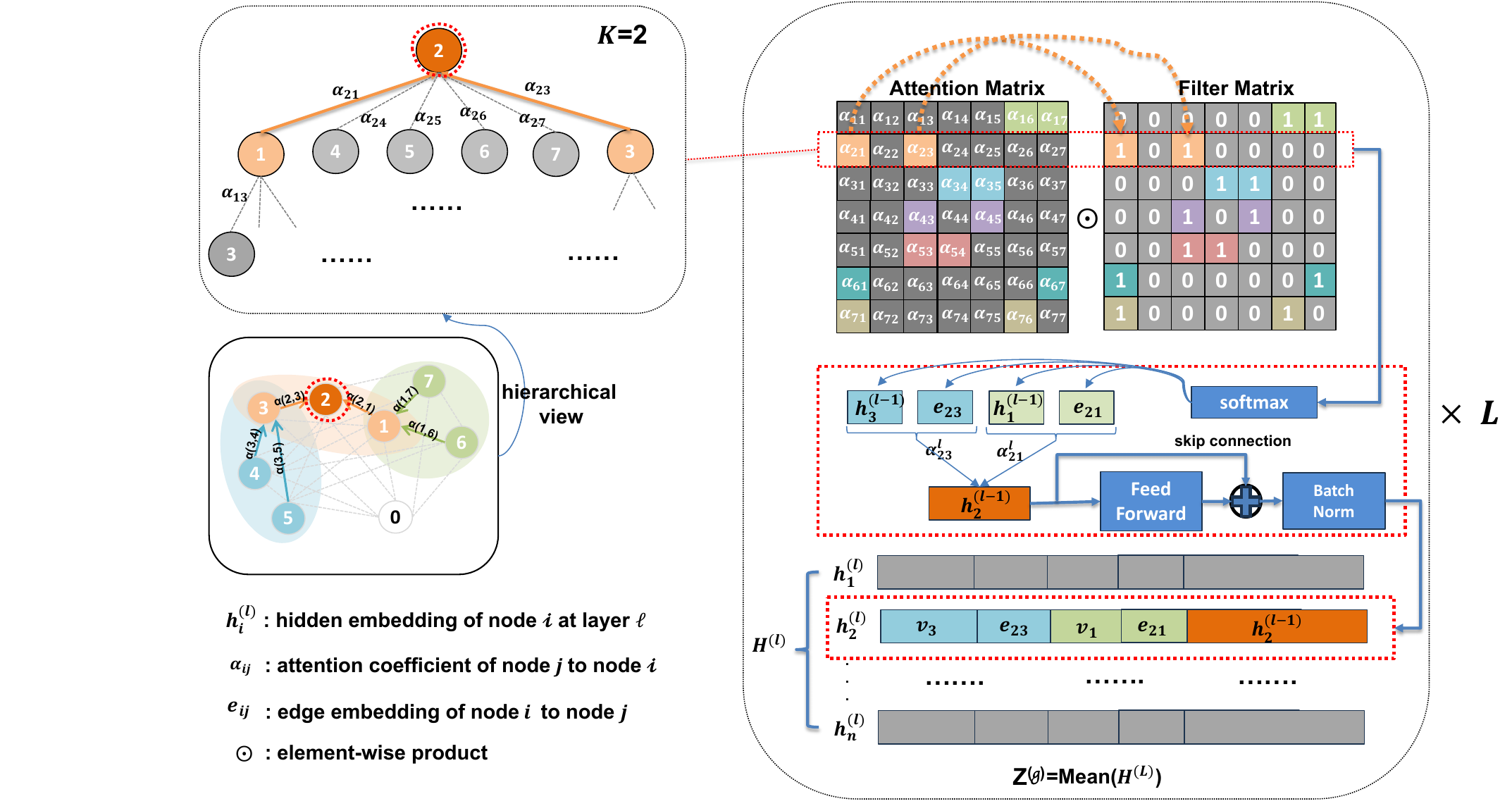}
\caption{Encoder with Attention Sampling, using node 2 as an example. The process involves sampling the top 2 nodes that are highly related to node 2 for aggregation. It's important to note that the attention matrix has rows representing the aggregating nodes, while the column elements represent the attention coefficients of the neighbouring nodes. The red box illustrates the operation process of a single node. All nodes execute their operations via matrix operations. The encoder is a residual network consisting of L layers. Each layer node combines the features of its top K neighbouring nodes and edges. The encoder's outputs are the hidden embedding $H^{(L)}$, which represents the node embedding, and the average value $Z(\mathbf{g})$, which shows the graph representation, that both are calculated after L layer residuals are computed.}
\label{figure2}
\end{figure*}

\subsection{Encoder with Attention Sampling}
We proposed a novel encoder (Fig. \ref{figure2}) that avoids drawing the full expansion of graph representation like Graph Convolution Network(GCN)\citep{kipfSemiSupervisedClassificationGraph2017a}, whose node aggregation weight is stable at each convolution layer, our strategy updates the node embedding by iteratively sample K high-related nodes that may affect the selection of decoder. Specifically, note an aggregating node set as $V_{agg}^l$, $V_{agg}^l \in V$ while the unduplicated neighbours of any node $i$ in set $V_{agg}^l$ are denoted as set $V_{i\_agg}^l$. Our encoder strategy first computes a correlation matrix $\hat{A}$ shown in Eq.(\ref{eq11}) by attention mechanism. $\hat{A^l}$ has the same dimension as the adjacency matrix $A$ without depot as it is selected when the vehicle load is full.

\begin{equation}\label{eq11}
    \hat{A^l} = \begin{bmatrix}
        \alpha_{11}^l & \cdots &\alpha_{1n}^l\\
        \vdots & \ddots & \vdots \\
        \alpha_{n1}^l & \cdots & \alpha_{nn}^l
    \end{bmatrix}
\end{equation}

Following the Transformer architecture \citep{vaswaniAttentionAllYou2023}, each element $\alpha_{i j}^{l}$ $\in [0,1]$ in $\hat{A^l}$ represents the attention coefficient that node $j$ to node $i$ at layer $l$, as shown in Eq.(\ref{eq12}).$W_q$ and $W_k$ are learnable matrices, and different learnable matrices with dimensions $d_q$ and $d_k$ are used at different attention layers to enhance the encoder's generative ability. $\sigma$ is softmax function.

\begin{equation}\label{eq12}
     \alpha_{i j}^{l} = \sigma\left( \frac{\left(W_q^l x_i^{(l-1)}\right)^T\left( W_k^l\left(x_j^{(l-1)}+e_{ij}^{(l-1)}\right) \right)}{\sqrt{d_k}} \right)
\end{equation}

The encoder then samples $K$ highly correlated neighbours for each node using a neighbour filter $N_f^l \in \mathcal{R}^{N\times N}$, where the $K$ neighbours are noted as 1, and 0 for those converse nodes, shown as Eq.(\ref{eq13}), and $\delta$ is the ranking function to identify the position of an element-wise flag to be 0 or 1.

\begin{equation}\label{eq13}
    N_f^l = \delta \left(\hat{A^l}\right)
\end{equation}

To select the attention coefficient of high-related nodes,   $\alpha$ of lower correlation nodes in the matrix $\hat{A^l}$ is set to be 0, achieved by performing the Hadamard product operation on the matrix $\hat{A^l}$ and $N_f^l$ at each layer $l$. 
When performing feature fusion, neighbour nodes $V_{i\_ne}^l$ with $\alpha$ value 0 will not contribute their characteristics to aggregating nodes $i$ while high-related nodes aggregate their feature through another learnable weight matrix $W_v^l$ and attention coefficient $\alpha$. The formula (\ref{eq14}) indicates the operations on matrices while the formula (\ref{hidden embedding element}) shows the calculation of hidden embedding in each layer from the element view. Here we still use $\sigma$ for softmax function and $\alpha_{ij}^\prime$ for exact elements of $\alpha_{ij}$ after softmax, that is $\alpha_{ij}^\prime = \sigma\left(\alpha_{ij}\right)$.

\begin{equation}\label{eq14}
    H^l =  BN\left(\left(W_v^lH^{(l-1)}\right)\cdot \sigma \left(\hat{A^l}\odot N_f^l\right)^T\oplus H^{(l-1)}\right)
\end{equation}

\begin{equation}\label{hidden embedding element}
    h_i^l = BN \left( \sum_{j=1}^{||V_{i\_agg}^l||} \left( \alpha_{ij}^{\prime (l-1)}w_v \left( h_j^{(l-1)} + e_{ij}^{(l-1)} \right) \right) \oplus h_i^{(l-1)}  \right)
\end{equation}

\begin{align}
    Z(\mathbf{g}) & = MEAN(MLP(H^{(L)} \oplus H^0))\\
    & = \bar{x}
\end{align}
    
Note that $\odot$ is Hadamard product operation. Each sublayer utilizes the skip-connection \citep{szegedyInceptionv4InceptionResNetImpact2017} $\oplus$ and batch normalization(BN) \citep{ioffe2015batch} to prevent vanishment. The value of K is a hyperparameter, and the selection of K affects the accuracy and generalization ability of the model to a certain extent. The encoder’s outputs are the hidden embedding $H^{(L)}$, which represents the node embedding, and the average value $Z(g)$, which shows the graph representation. As the graph embedding will be an important context input during decoding, instead of solely relying on the mean of all node embeddings as the full graph representation, like \citep{koolAttentionLearnSolve2019,leiSolveRoutingProblems2022}, we enhance the process by initially leveraging the initial node vector $H^0$ and the encoder's output embedding vector $H^L$ to execute a concatenation operation. Subsequently, we feed it into a shallow multi-layer perceptron (MLP) and ultimately derive the final full graph representation \citep{zhangGPSPolicydrivenSampling2022} through a row-wise averaging operation. To streamline notation, $ \bar x $ is denoted to represent the graph-level readout.

\subsection{Decoder}
The decoder will generate a feasible solution sequentially, following a similar approach to \citep{koolAttentionLearnSolve2019} and \citep{leiSolveRoutingProblems2022}. During the decoding process, shown in Fig. \ref{figure1}(c), the Multi-Head Attention (MHA) \citep{vaswaniAttentionAllYou2023} mechanism takes a context embedding as its query input at each time step. The context vector at the initial time step comprises the full graph representation $\bar x$ and two node feature vectors. These node feature vectors represent the output of the first time step $ h_0^{L} $ and the output of the previous time step $h_{\pi_{t-1}} $, respectively. The context vector is shown in Eq.(\ref{eq context1}) and we use $W_{c}$ to process the linear projection.

\begin{equation} \label{eq context1}
    c_t^m = \left\{ \begin{array}{ll}
         W_{c^\prime}\left( \bar x || h_0^L  || Q_t \right) & t = 1\\
         \\
         W_{c^\prime}\left( \bar x  || h_{t-1}^L || Q_t \right) & t > 1\\
         
    \end{array}\right. 
\end{equation}
Here the superscript $m$ means MHA layer. We use the remaining vehicle capacity $Q_t$ as well to be part of context embedding at MHA layer. To ensure the capacity constraints, the remaining demand of nodes $q_{i,t}$ and the vehicle $Q_t$ is tracked at each time step as below:

\begin{equation}
    q_{i,t+1} = \left\{ \begin{array}{ll}
         q_{i} & \pi_t \neq i \\
        \\
        0 & \pi_t = i \\
         
    \end{array}\right. 
\end{equation}

\begin{equation}
    Q_{t+1} = \left\{ \begin{array}{ll}
         Q & \pi_t = 0 \\
         \\
         max(0,  (Q_t - q_{\pi_t,t})) & \pi_t \neq 0\\
         
    \end{array}\right. 
\end{equation}

We first define three weight matrices that can be learned through training to calculate the MHA value. These matrices are denoted as  $W_{q^\prime} \in \mathbb{R}^{d_h\times d_x}$, $W_{k^\prime} \in \mathbb{R}^{d_h\times d_x}$, and $W_{v^\prime} \in \mathbb{R}^{d_h\times d_x}$. Here, $d_x$ represents the dimension of encoder node embedding $h_i^L$, and $d_h$ is defined as $\frac{d_x}{H}$. The value of H, which represents the number of heads used to determine $d_h$, is also used to compute the MHA value through a diffusion aggregation method. The resulting context value is denoted as $c_t^s$ and recorded in the middle layer. The superscript $s$ means a single attention layer, that is, it will be the context query of the next single-head attention layer. Similar to our encoder, we use Eq.(\ref{head-coe-cus}) to calculate the attention coefficient for query node $c_t^m$ with key node $h_i^L ,i \in \{0,1,...,n\}$ at each time step $t$. 

\begin{equation}\label{head-coe-cus}
    u_{i,t}^m = \begin{array}{ll}
      \frac{\left(W_{q^\prime}c_t^m\right)^T
        \left(W_{k^\prime}h_i^L \right)}{\sqrt{d_h}}   & i \neq \pi_{t^\prime} , \forall t^\prime < t\\
         \\
    \end{array} 
\end{equation}

\begin{equation}\label{eq-multihead}
    c_t^s = W_m\left(||_{h^\prime =1}^H\sum_{i=1}^n\sigma\left(u_{i,t}^m\right)^{h^\prime}\left(W_{v\prime}h_i^L\right)^{h^\prime}\right)
\end{equation}
The formula (\ref{eq-multihead}) illustrates the process of multi-head attention parallel aggregation to generate single-head attention layer context input $c_t^s$, where $||$ is a serial concatenation operation. $H$ is the number of attention heads, equivalent to the node embedding vector divided into $H$ parts. A reminder here is that $h^\prime$ is the head number while $h$ is the hidden embedding of our encoder. The context node $c_t^m$ performs attention calculations with each part of the divided node embedding vector and then merges the results to enhance the generalization of the attention mechanism to the perception of different feature areas. 

The single-head attention layer coefficient $u_{i,t}^s$ can be calculated using the formula (\ref{ctanh}) with $c_t^s$ and each node embedding $h_i^L$. 
\begin{equation}\label{ctanh}
    u_{i,t}^s = C \cdot tanh\left( \frac{(W_{q^{\prime\prime}} c_t^s)^T(W_{k^{\prime\prime}} h_i^L)}{\sqrt{d_x}} \right)
\end{equation}
Motivated by work \citep{belloNeuralCombinatorialOptimization2017, koolAttentionLearnSolve2019, leiSolveRoutingProblems2022}, here we first control the attention coefficient within an interval $[-C,C]$ through $tanh$ and parameter $C$, with the purpose of increasing the distinction of each node. When given the current state of the GASE model $G$ and the previous solution nodes $\pi_{t^\prime}$, this result is then used to calculate the probability distribution of all nodes to be output by the decoder at the current time step through softmax function $\sigma$ according to Eq.(\ref{prodis}). 
\begin{equation}\label{prodis}
     p_\theta\left(\pi_t \mid S, \pi_{t^\prime}, \forall t^\prime < t \right) = \sigma(u_{i,t}^s)
\end{equation}

Then decoder samples according to this probability distribution with a stochastic property at each time step to form a feasible solution. 

However, it's important to note that the attention mechanism alone cannot guarantee that the output node satisfies the constraints. Therefore, this article utilizes masking technology to ensure that the output node of each time step meets the problem constraints. Attention, We employ distinct masking techniques \textbf{before the softmax} \citep{feyFastGraphRepresentation2019} function at both multi and single attention layers to address the constraints of customer nodes and the depot. For customer nodes, those that have already been visited and nodes with demands that exceed the current vehicle capacity are inaccessible, and their attention coefficients are set to negative infinity. That is, $u_{t,i}^m, u_{t,i}^s = -\infty,$ if $\{\left(i\neq 0, i\in \pi_t^\prime, \forall t^\prime < t\right)$ or $ \left( q_i > Q_t \right)\}$. As a result, when calculating the access probability of each node at the current time step, the access probability of the inaccessible node will be 0. Furthermore, the depot node cannot be visited twice consecutively within a sub-path. This means that when $t=1$ or $\pi_{t-1} = 0$, the depot will be masked at $t$ as well, which can be represented as $u_{t,i}^m, u_{t,i}^s = -\infty,\{\left(i=0, t =1\right)$ or $\left(i=0,\pi_{t-1} = 0\right)\}$. Moreover, once the depot is accessed, the remaining capacity of the vehicle will return to the maximum vehicle capacity by dynamic updating $Q_t = Q, \ if\  \pi_t = 0$.

\subsection{Training}
This paper utilizes DRL schema to train the encoder-decoder GASE model. Reviewing the GASE process, given an instance, a sequential feasible node solution $\pi$ is used to calculate the reward in a DRL schema through Eq. (\ref{eq7}) with a probability $p_\theta(\pi\mid G)$ in Eq. (\ref{eq8}). According to Monte Carlo estimation, the loss function $\mathcal{L}(\theta\mid G)$ of model training can be defined as the expected solution reward under the current learnable parameters $\theta$ and an instance graph $G$, i.e. $\mathcal{L}(\theta\mid G) = \mathbb E_{\pi\sim p_\theta(\pi\mid G)}[L(\pi)] $. To enhance the model performance and convergence speed, we choose the Actor-Critic \citep{mnihAsynchronousMethodsDeep} method and baseline REINFORCE algorithm \citep{suttonPolicyGradientMethods1999}. More specifically, the GASE model, with parameter $ \theta $, incorporates both actor and critic policy networks but differs in its solution node sampling approach. The actor policy network, denoted as $ \pi _ \theta $, utilizes a stochastic sampling strategy and is reinforced by the baseline critic network, $ \pi _ \theta ^b $, which generates a greedy rollout. That means when the decoder outputs the solution node, the baseline (critic network) always selects the node with the highest probability after the softmax, while the actor model is sampled according to the probability distribution. The policy gradient is shown as:

\begin{equation}
    \nabla \mathcal{L} \left( \theta | G \right) = \mathbb E_{\pi\sim p_\theta(\pi | G)}[(L(\pi | G) - b) \nabla_\theta log p_\theta(\pi | G)]
\end{equation}
where $ b $ represents the length of a solution generated by the critic policy network using a greedy decoding strategy. 

This improved actor-critic structure utilizes self-critic models, the critic network outperforming randomly sampled actors in the initial stage of training due to the characteristics of greedy reward. However, it's important to note that relying on a greedy model can result in easily reaching local optimal solutions. Consequently, it will continuously update the critic network's parameters to ensure the baseline rollout remains the most optimal network model. If the difference between the GASE model and the baseline model, aka the REINFORCE, denoted as $(L( \pi | G) - b)$, is negative, it indicates that the current GASE model (i.e., the actor network) outperforms the critic baseline rollout. In such cases, it is essential to consider updating the critic network to align with the current actor network in the next training iteration. To ensure effectiveness, the baseline is only updated when the difference in path length calculated between the actor-network and the baseline on the validation set meets the significance level of the paired t-test, specifically when its significance value is less than 5\%. Updating the enhanced baseline will further improve the performance of the actor network until the policy gradient value converges or all training epochs are completed. The baseline REINFORCE algorithm is illustrated in Algorithm \ref{alg1}.

\begin{algorithm}
	\renewcommand{\algorithmicrequire}{\textbf{Input:}}
	\renewcommand{\algorithmicensure}{\textbf{Output:}}
    \caption{Baseline REINFORCE Algorithm}\label{alg1}
\begin{algorithmic}[1]
    \REQUIRE Actor policy network $\pi_\theta$; Baseline policy network $\pi_\theta^b$; Training epochs $E$; Batch size $B$; Steps per epoch $T$; Significance level $\alpha$.
    \STATE Initialization: $\theta, \theta^b \gets$ Xavier init $\theta$;
    \FOR{ $e$ in $1...E$}
    \FOR{ $t$ in $1...T$}
    \STATE $G_i \gets RandomInstance(), \forall i \in \{1,....,B\}$
    \STATE $\pi_i \gets SampleSolution(G_i, p_\theta), \forall i \in \{1,....,B\}$
    \STATE $\pi_i^b \gets GreedyRollout(G_i, p_\theta^b), \forall i \in \{1,....,B\}$
    \STATE Compute $L(\pi_i | Gi), L(\pi_i^b | Gi) $ through Eq. (\ref{eq7}) 
    \STATE $\nabla\mathcal{L} \gets \sum_{i=1}^B (L(\pi_i) - L(\pi_i^b))\nabla_\theta log p_\theta(\pi_i | G_i)$
    \STATE $\theta \gets Adam(\theta, \nabla_\theta\mathcal{L})$
    \ENDFOR
    \IF{OneSidedPairedTTest($p_\theta, p_\theta^b < \alpha$)}
    \STATE $\theta^b \gets \theta$
    \ENDIF
    \ENDFOR
    \ENSURE Parameters set $\theta$ of actor network

\end{algorithmic}  
\end{algorithm}

\section{Experiments}
We conducted a series of experiments to assess the effectiveness of our proposed GASE model. We evaluated the model's ability to generate optimal length routes, the speed at which it performs training and inference, its generalization capability, and the sensitivity of its parameters. The experimental results were analyzed using both real and simulated data sets to address the following four research questions:
Q1 (Solution Performance): Does the graph attention-based sampling model we propose have a competitive advantage over other state-of-the-art models in terms of the length of its generation path, the speed of its training and inference, and other factors? Are there any advantages over other models?
Q2 (Sensitivity Analysis): How does the performance of the GASE model change when we modify the sampling parameter K?
Q3 (Generalization Ability): As pure data-driven models often depend on the distribution of data, does the GASE model have the ability to solve problems of different scales and demonstrate generalization ability in solving medium- and large-scale problems compared to other data-driven models?
\subsection{Experiment Setup}
\subsubsection{Dataset}
When addressing the VRP, previous studies have shown a tendency to process data sets in a similar manner. These data sets are typically randomly generated or consist of simulated or real data, such as those found in CVRPLIB \cite{uchoa2017new}, to assess the model's performance. This paper aims to simultaneously evaluate both the simulated and real data sets, comprehensively analysing the model's accuracy, computational speed, and generalization ability. Similar to prior research, this paper uses different random seeds to generate node coordinates and requirements of varying scales for the simulation data set. Additionally, the performance of our proposed GASE model is evaluated using CVRPLIB dataset.

\subsubsection{Baseline}
When selecting a baseline, we chose the optimal solution generator Gurobi \citep{Gurobi2024}, the popular heuristic algorithm LKH \citep{linEffectiveHeuristicAlgorithm1973}, and the solver OR-tools \citep{ortools} developed by Google for comparison. For the data-driven model based on reinforcement learning, we selected three state-of-the-art models that applied different popular deep learning models in different periods: the long short-term memory (LSTM) network called pointer network sequence model (PtrNet) \citep{vinyalsPointerNetworks2017}, the attention mechanism sequence model based on the transformer architecture (AM Model) \citep{koolAttentionLearnSolve2019}, and the graph model of graph attention residual network (E-GAT) \citep{leiSolveRoutingProblems2022}. These were chosen as the end-to-end learning-based baseline models to compare with GASE. The first two models were prevalent, and the AM model was used as the basic model for many works. The third graph model claims to have achieved the latest state-of-the-art results.

\subsubsection{Parameter settings}
To ensure the reliability of experimental results, we have carefully maintained certain basic parameters for the problem and neural network model. These include random seed settings, maximum vehicle capacity under different problem scales, number of layers for graph neural network aggregation, neural network optimizer, parameter initialization settings, reinforcement learning training settings, and more. The specific software and hardware parameter settings can be found in the table \ref{tabparamter}. Additional hyperparameters of the model proposed in this article will be discussed in detail in the sensitivity analysis section.
\begin{table}[th!]
\caption{Parameter Settings}
\label{tabparamter}
\centering
\begin{tabular}{ll}
\toprule
Parameters & Setting\\
\hline
GPU & Nvidia RTX A6000 \\

CPU &  Intel Xeon Silver 4126 2.10GHz \\

Vehicle Max Capacity & \{node 20:30; node 50: 40; node 100: 50\} \\

customer need & random from 1 to 9\\

Optimizer &  Adam \citep{kingma2014adam}\\

Network parameter initialize & Xavier \citep{glorot2010understanding}\\

Training instances & 128000\\

Training epoch & 200\\

batch size & 128\\

Learning rate &  $3 \times 10^{-4}$ \\

Learning rate decay $\beta$ & 0.96\\

encoder layers & 4\\

\bottomrule
\end{tabular}
\end{table}

\subsection{Evaluation Results}

The model's performance is primarily assessed based on the average length of the solution on 1000 instances of the test set. To evaluate the quality of the model, we compare the path length and inference speed of each model on the test set, while keeping the random seeds the same. During testing, we consistently apply the greedy decoding method across the test set. Our experiments have shown that, when the parameter model of the neural network is fixed, the output results of the greedy strategy generally outperform the sampling strategy. This finding aligns with the AC algorithm used during training, where Greedy (critic) guides Sampling (actor). GASE performance results are presented in Table \ref{basicresult} and Table \ref{vrplibresult}. The gap indicator refers to the difference between the current model and the best baseline model on the test set.
\subsubsection{Solution Performance for Q1 }
Based on table \ref{basicresult}, our proposed model has improved the quality of solutions in VRP of different scales under similar conditions, also accelerating the inference time in comparison to graph models. Compared to other models, the gaps between our model and the current best baseline are only 1.80\%  on VRP20, 3.46\%  on VRP50, and 4.60\% on VRP100, which outperformed other deep learning-based models. This result demonstrates that based on attention sampling, our proposed graph encoder is better at extracting the features of nodes and edges in VRP. This allows for the output of better node representation and graph representation as input to the decoder during the decision-making process. Our model is less affected by poor nodes and edges, which leads to an improvement in the quality of the solution. Additionally, our matrix-based filter parallelizes the sampling calculation and encoding process, making full use of the parallel computing characteristics of the GPU. This has reduced the training time and inference time, making the model easier to converge. The AM model and E-GAT model employ a sampling search mechanism that exhaustively seeks optimal solutions within a constrained range. This approach, however, incurs high time complexity and results in prolonged inference times, ranging from several minutes to several hours depending on the problem size. Consequently, this diminishes the efficiency and timeliness of problem-solving. In contrast, our model strikes a superior balance, ensuring high-quality results while significantly enhancing inference speed. This comprehensive comparison underscores our model as the more effective choice for timely and efficient problem resolution.

\begin{table*}[h!]
	\centering
	\caption{compare with other approaches}
        \label{basicresult}
       \resizebox{\linewidth}{!}{
	\begin{tabular}{cccccccccc}
		\toprule
		\multirow{2}*{Model} & \multicolumn{3}{c}{CVRP20} &  \multicolumn{3}{c}{CVRP50}& \multicolumn{3}{c}{CVRP100}\\ 
		\cmidrule{2-4}\cmidrule{5-7}\cmidrule{8-10}
        \noalign{\smallskip} 
		& Length & Gap(\%) & Time & Length & Gap(\%) & Time & Length & Gap(\%) & Time\\
		\hline 
		Gurobi       & 6.10 & 0.00 & -  & -     & -    & -  &   -   &  -   &  -  \\
		LKH          & 6.14 & 0.58 & 2h & 10.38 & 0.00 & 7h & 15.65 & 0.00 & 13h  \\
		OR Tools     & 6.43 & 5.41 & -  & 11.31 & 9.01 & -  & 17.16 & 9.67 &  -  \\
		\cdashline{1-10}[0.8pt/2pt]
            PtrNet       & 6.59 & 8.03 & 0.11s  & 11.39 & 9.78 & 0.16s  & 17.23 & 10.12 & 0.32s \\
            PtrNet(bs)   & 6.40 & 4.92 & 0.16s  & 11.15 & 7.46 & -  & 16.96 & 8.39 & -  \\
		AM model     & 6.4 & 4.97 & 1s  & 10.98 & 5.86 & 3s & 16.8 & 7.34  & 8s  \\
            AM model(sampling)  & 6.25 & 2.49 & 6m  & 10.62 & 2.40 & 28m & 16.23 & 3.72  & 2h \\
		E-GAT        & 6.26 & 2.60 & 2s & 10.80  & 4.05 & 7s & 16.69 & 6.68 & 17s \\
            E-GAT(sampling)     & 6.19 & 1.47 & 14m & 10.50  & 1.54 & 1h & 16.16 & 3.25 & 4h \\
            \cdashline{1-10}[0.8pt/2pt]
            
		{\bf Our approach}  & {\bf 6.21} & {\bf 1.80} & {\bf 1.7s} & {\bf 10.74} & {\bf 3.46} & {\bf 2.98s} & {\bf 16.37} & {\bf 4.60} & {\bf 6.65s} \\
		\bottomrule
	\end{tabular}
 }
\end{table*}

\subsubsection{Sensitivity Analysis for Q2}
\begin{figure}[h!]
    \centering
    \subfigure[Convergence Curve of the GASE Model on CVRP20 with Different Sampling Rates Using a 4-Layer Residual Connection and 8-Head Attention Mechanism] {\includegraphics[width=.48\textwidth]  {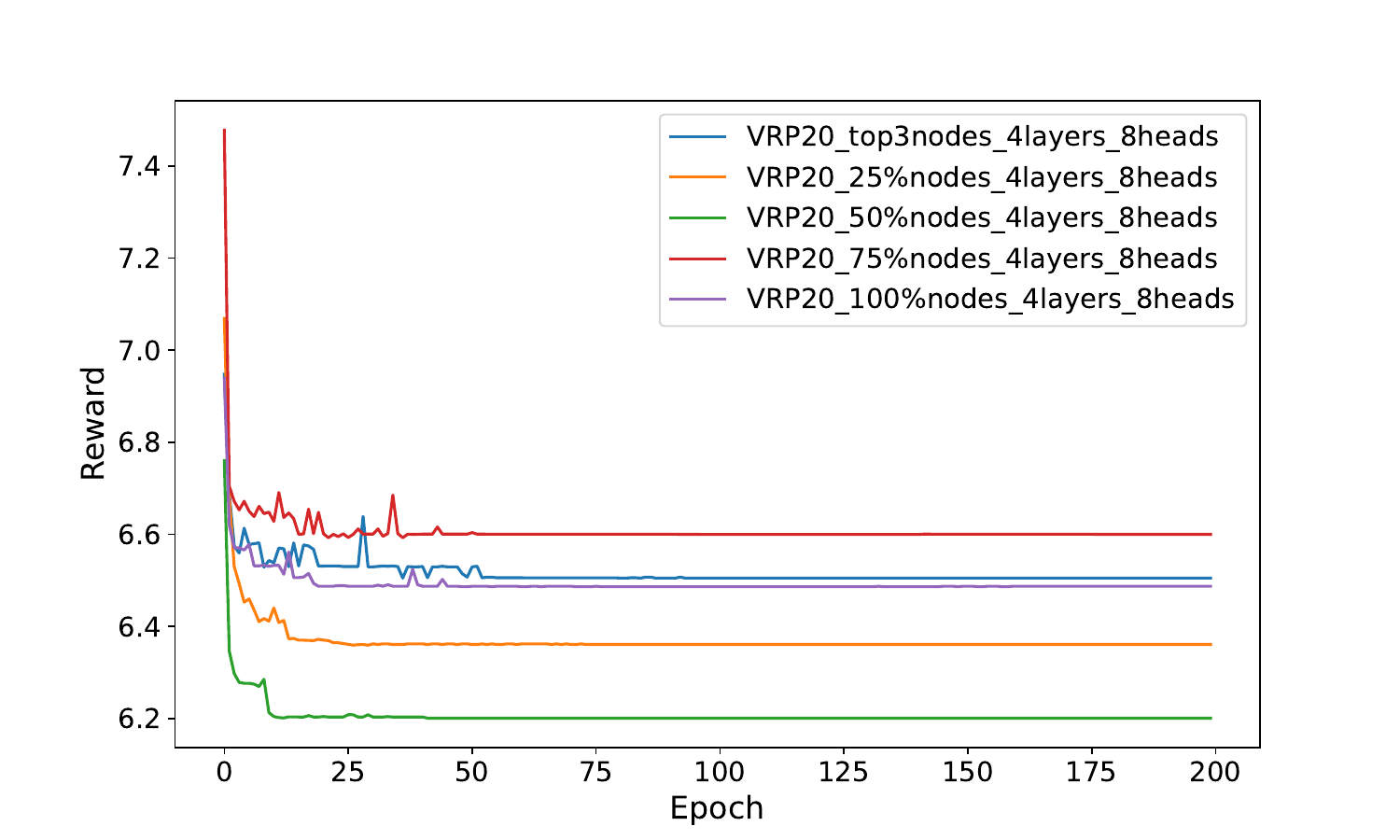}}
    \subfigure[Convergence Curve of the GASE Model on CVRP50 with Different Sampling Rates Using a 4-Layer Residual Connection and 8-Head Attention Mechanism] {\includegraphics[width=.48\textwidth]{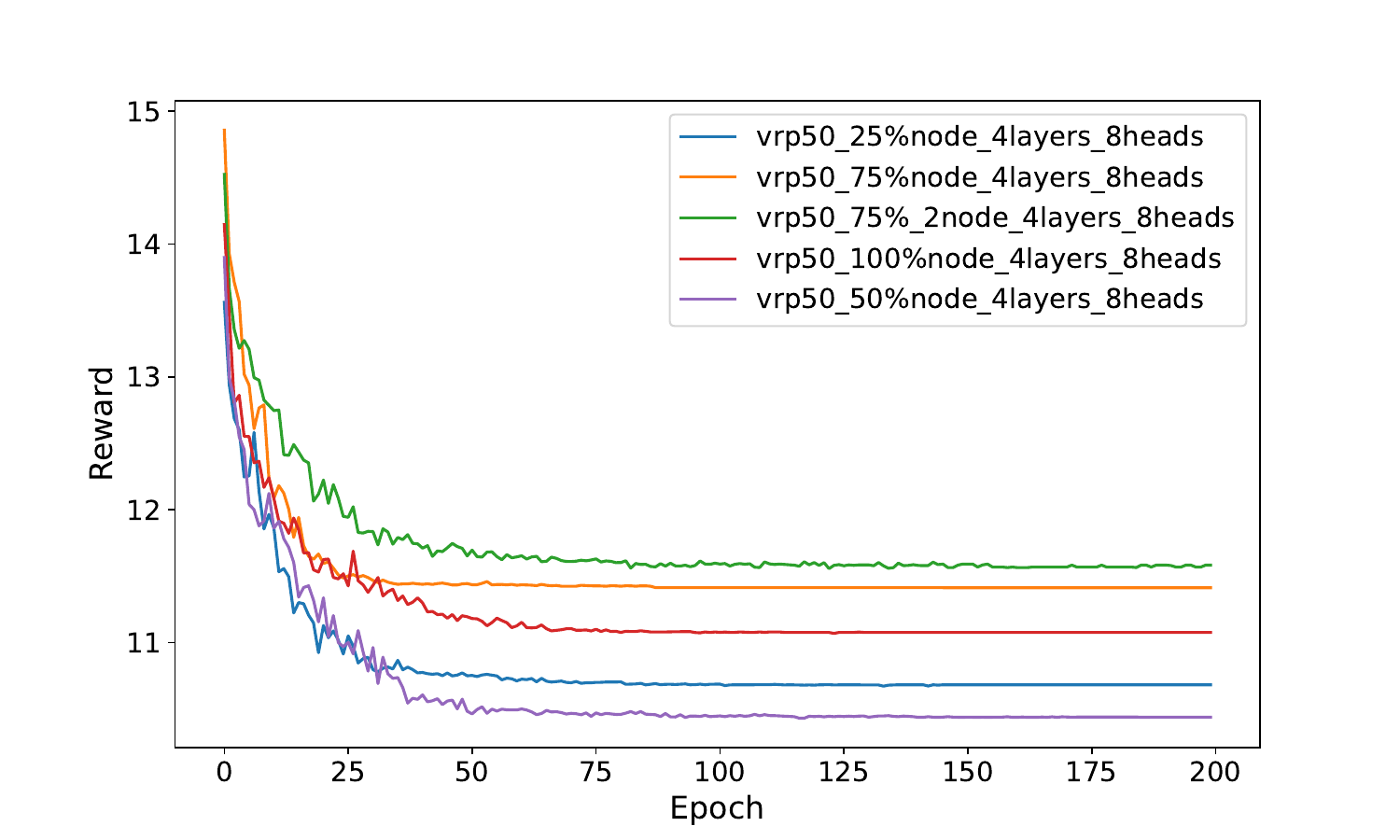}}
\caption{Validation performance of GASE for several skip-connection layers on  problem size 20/50 and different sampling rate $K$.}
\label{fig_Sensi_samples}
\end{figure}

\begin{figure*}[h!]
    \centering
    \subfigure[Convergence Curve of the GASE Model on CVRP20 with 2-Layer Residual Connection] {\includegraphics[width=.48\textwidth]  {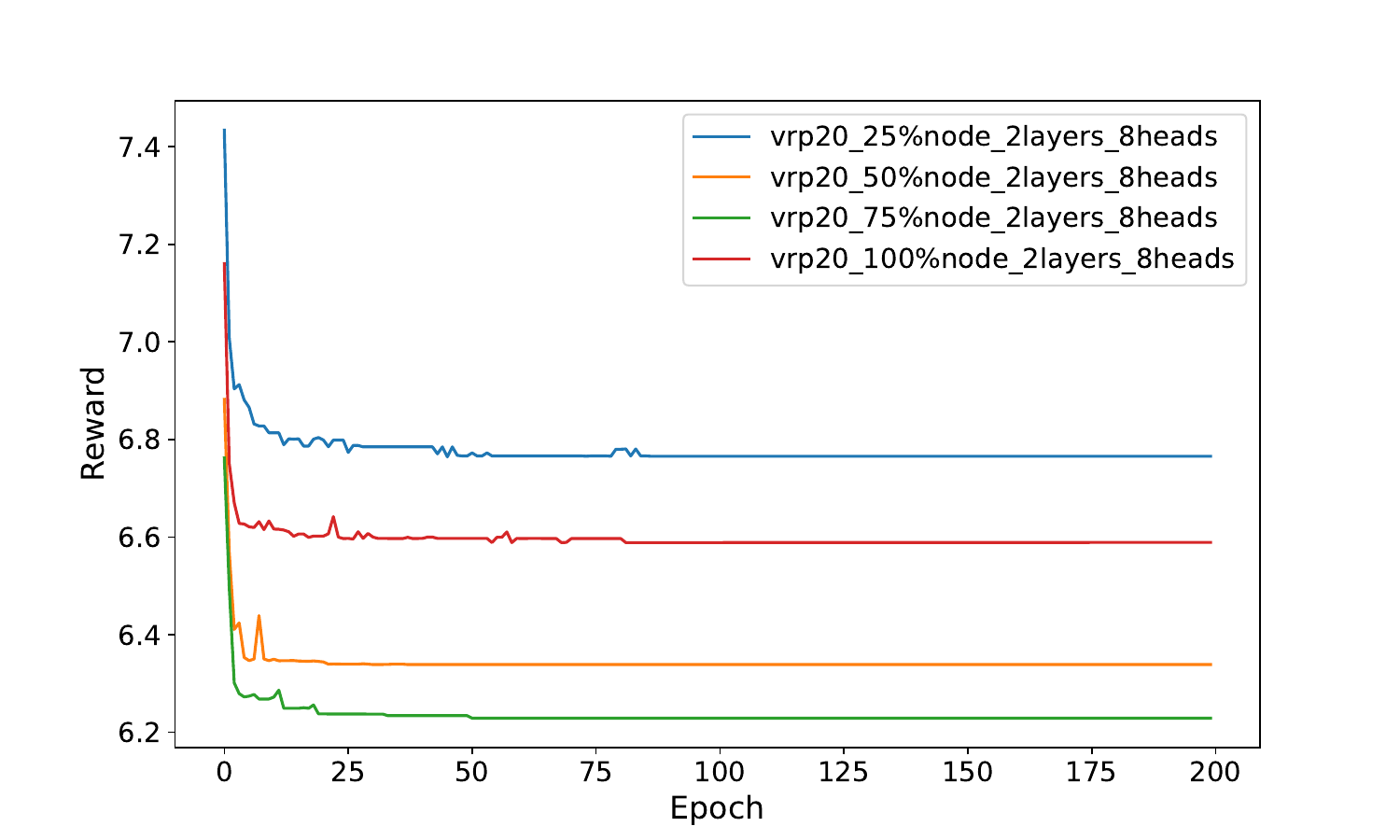}}
    \subfigure[Convergence Curve of the GASE Model on CVRP20 with 3-Layer Residual Connection ] {\includegraphics[width=.48\textwidth]{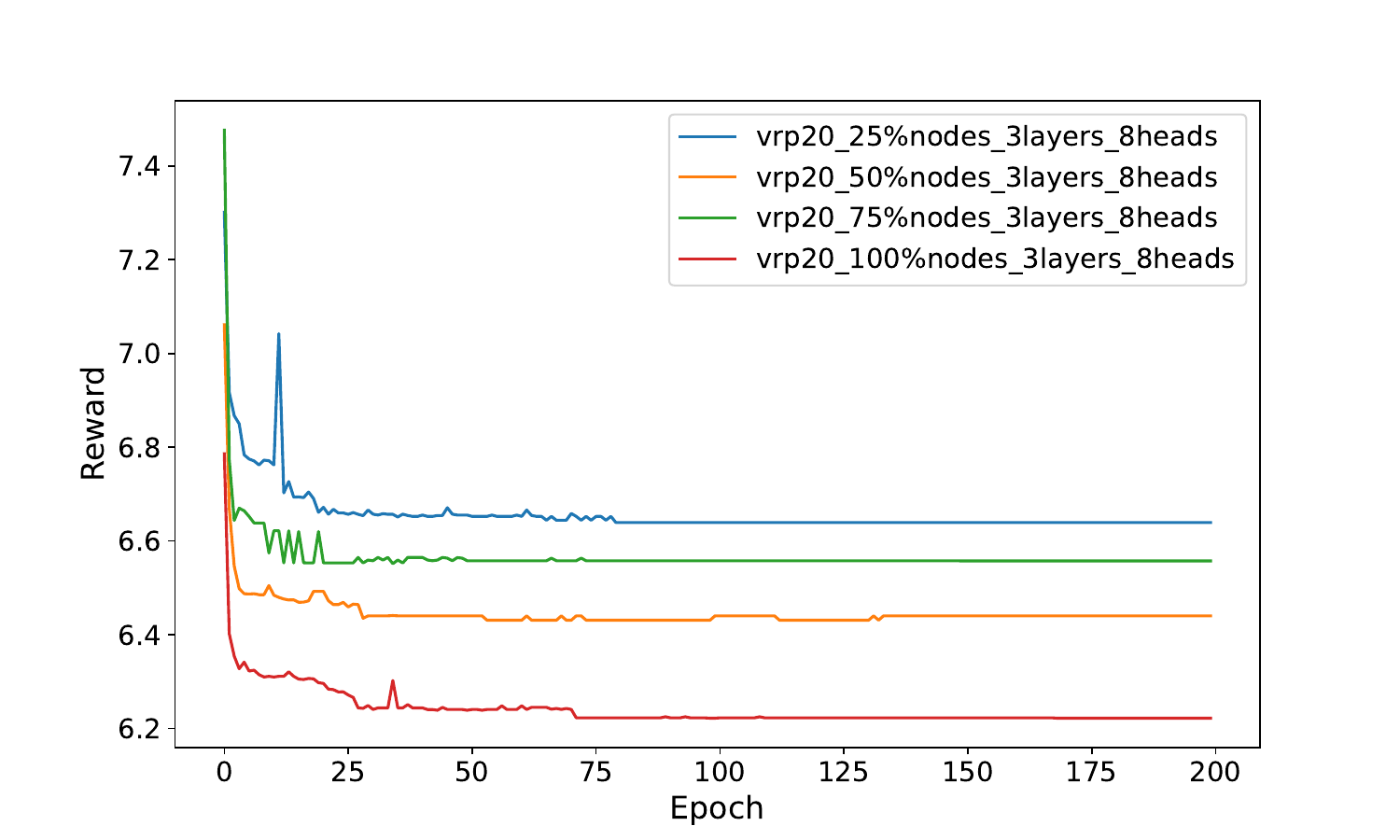}}
    \subfigure[Convergence Curve of the GASE Model on CVRP20 with 4-Layer Residual Connection] {\includegraphics[width=.48\textwidth]{sensitive_pics/vrp20_4layers_8heads.pdf}}
    \subfigure[Convergence Curve of the GASE Model on CVRP20 with 5-Layer Residual Connection] {\includegraphics[width=.48\textwidth]{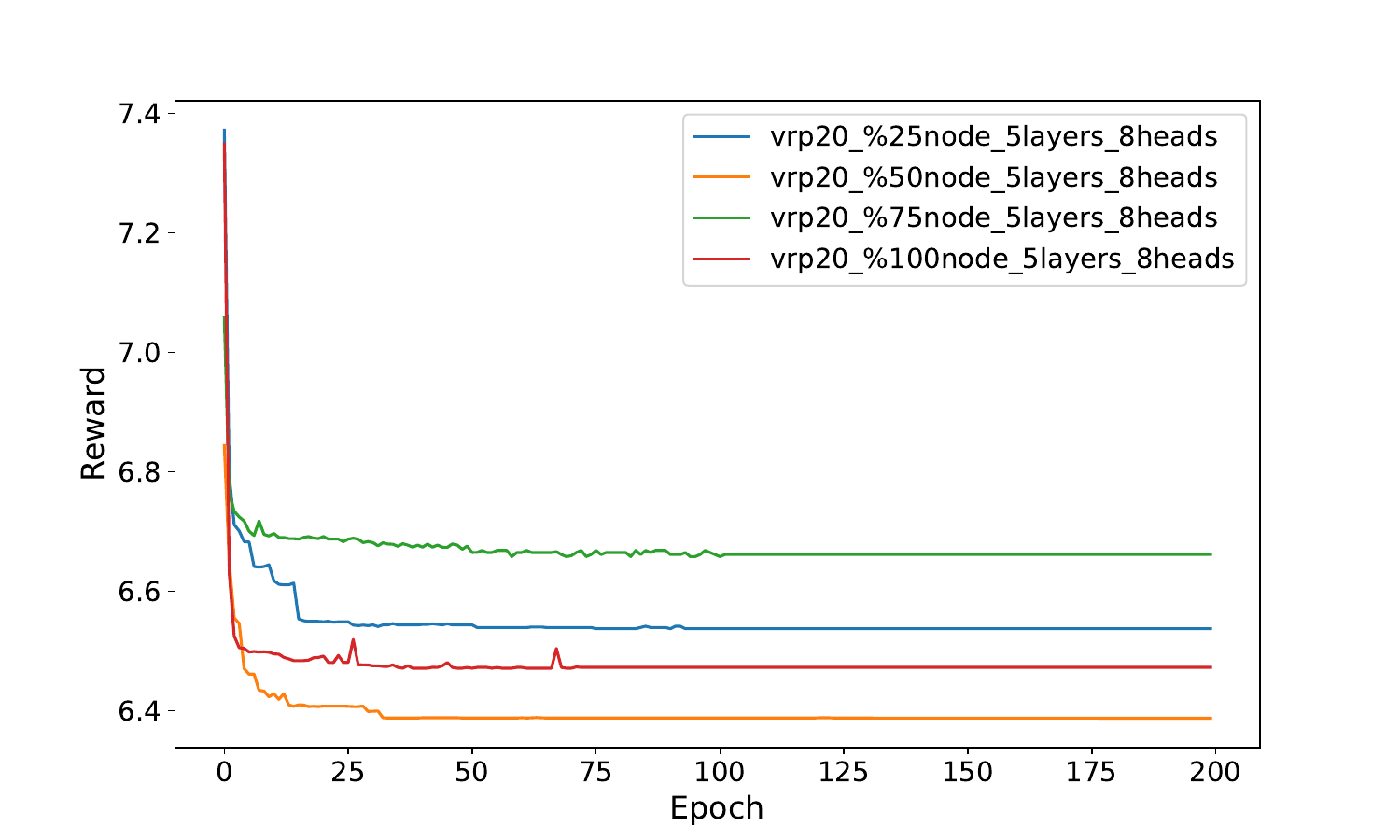}}
\caption{Validation performance of GASE for different skip-connection layers on  problem size 20 and various sampling rate $K$.}
\label{fig_Sensi_layers}
\end{figure*}

\begin{figure*}[h!]
    \centering
    \subfigure[Convergence Curve of the GASE Model on CVRP20 with Single-Head Attention] {\includegraphics[width=.49\textwidth]  {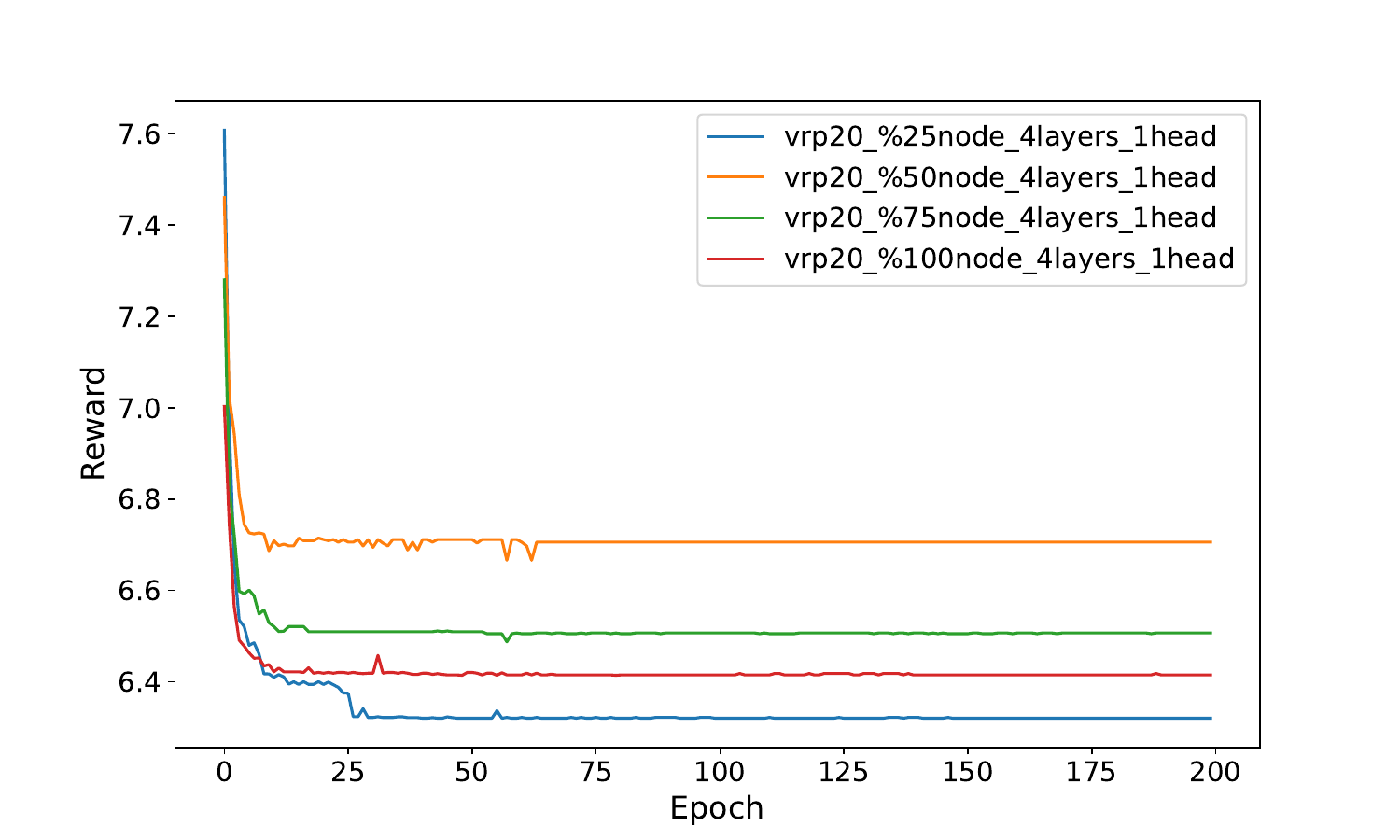}}
    \subfigure[Convergence Curve of the GASE Model on CVRP20 with 4-Head Attention] {\includegraphics[width=.49\textwidth]{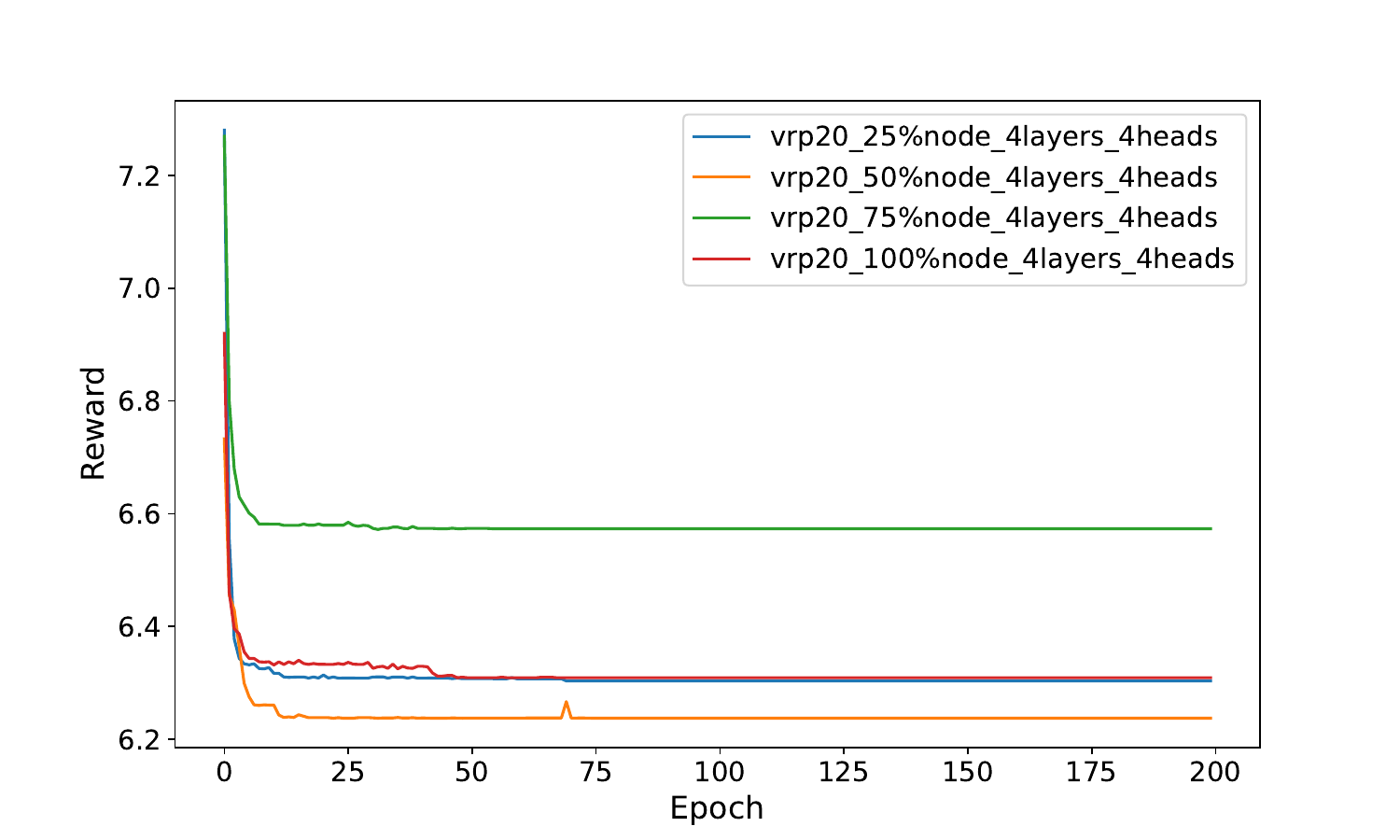}}
    \subfigure[Convergence Curve of the GASE Model on CVRP20 with 8-Head Attention Mechanism] {\includegraphics[width=.49\textwidth]{sensitive_pics/vrp20_4layers_8heads.pdf}}
\caption{Validation performance of GASE for different attention heads on  problem size 20 and various sampling rate $K$.}
\label{fig_Sensi_heads}
\end{figure*}

We investigated the impact of key parameters in the GASE model, including sampling size, number of attention heads, and skip-connection layers, as shown in Fig. \ref{fig_Sensi_samples}, Fig. \ref{fig_Sensi_layers} and Fig. \ref{fig_Sensi_heads}. In general, we fixed two of the hyperparameters and varied the remaining one to explore the model sensitivity.

 \textbf{Impact of sampling size:} The impact of varying sampling sizes on the performance of the proposed model was systematically examined across CVRP20 and CVRP50 problem instances. Throughout the evaluation, a fixed configuration was maintained, with the skip-connection layer set to 4 and the number of multi-heads set to 8 (will be analyzed in the following sections). The findings in Fig.\ref{fig_Sensi_samples} revealed that the model exhibited optimal convergence speed and performance metrics when a sampling size equivalent to 50\% of the problem scale was employed for both CVRP20 and CVRP50 instances. Consequently, a sampling strategy involving 50\% of the nodes has been adopted for problem instances of size 100. This decision is driven by the significant computational demands associated with training on larger problem scales.

\textbf{Impact of skip-connection layer numbers:} Our experimental findings, shown in Fig.\ref{fig_Sensi_layers}, underscore the critical influence of the number of layers utilized for integrating residual connections within the graph encoder network on the model's performance. Notably, an inadequate or excessive number of layers can detrimentally affect the model's efficacy across validation and test datasets. This phenomenon likely arises from the potential for extensive residual layers to induce over-smoothing during the graph feature encoding process, whereas a paucity of layers may inadequately capture feature embeddings. According to the findings of our empirical investigation, the effect of selecting the residual connection of a 4-layer encoder architecture is generally superior to that of the residual connections.

\textbf{Impact of the number of multi-head:} We note a discernible enhancement in the performance of our model with an increase in the number of attention heads, shown in Fig. \ref{fig_Sensi_heads}. This augmentation is attributed to the capacity of additional heads to yield more balanced and precise outcomes. However, it is imperative to acknowledge that such augmentation also extends the training and inference times. Building upon the insights gleaned from our observational experimentation, particularly with regards to the determination of optimal residual layers and sampling coefficients, we have opted to employ 8 attention heads. This decision is geared towards addressing larger vehicle routing problems with enhanced efficiency and effectiveness.

\subsubsection{Generalization Ability for Q3}
In this section, we investigate the impact of model generalization, specifically evaluating the model's performance across varying problem sizes. To this end, we extend our analysis to real-world data by employing the GASE model configuration on benchmark datasets from CVRPLIB. Specifically, we use a model trained on 50-node instances for problems with fewer than 50 nodes and the 100-node model for problems exceeding 50 nodes. The comparative results, shown in Table \ref{vrplibresult}, indicate that the GASE model demonstrates strong performance on real-world datasets with a lower average gap for uniformly distributed data. 

\begin{table*}[h!]  
	\centering
	\caption{Different Model Performance on CVRPLIB Datasets}
        \label{vrplibresult}
        \resizebox{\linewidth}{!}{
	\begin{tabular}{lcccccccc}
		\toprule
		\multirow{2}*{Data instance} & \multirow{2}*{Nodes number}& \multicolumn{1}{c}{Optimal Solution} &  \multicolumn{2}{c}{Ours Approach}& \multicolumn{2}{c}{E-GAT}&\multicolumn{2}{c}{AM Model}\\ 
		\cmidrule{3-3}\cmidrule{4-5}\cmidrule{6-7}\cmidrule{8-9}
        \noalign{\smallskip} 
		& & Length & Length & Gap(\%) & Length & Gap(\%) &  Length & Gap(\%)    \\
		\hline 
		A-n32-k5  & 31  & 784   & 827   & 5.48   & \textbf{789}     & \textbf{0.63}  & 839   & 7.01 \\
		A-n36-k5  & 35  & 799   & \textbf{805}   & \textbf{0.75}   & 839     & 5.00  & 878   & 9.89 \\
            A-n37-k5  & 36  & 669   & \textbf{693}   & \textbf{3.58}   & 710     & 6.12  & 711   & 6.28 \\
		A-n38-k5  & 37  & 730   & \textbf{735}   & \textbf{0.68}   & 751     & 2.87  & 762   & 4.38 \\
            A-n39-k5  & 38  & 822   & \textbf{831}   & \textbf{1.09}   & 838     & 1.94  & 840   & 2.19 \\ 
            A-n44-k6  & 43  & 937   & \textbf{950}  & \textbf{1.39}   & 984     & 5.01  & 997   & 6.40 \\
		A-n45-k6  & 44  & 944   & \textbf{963}  & \textbf{2.01}   & 984     & 4.23  & 1015  & 7.52 \\
            A-n46-k7  & 45  & 914   & 1066  & 16.63  & 999     & 9.29  & \textbf{997}   & \textbf{9.08} \\
            A-n48-k7  & 47  & 1073  & 1127  & 5.03    & \textbf{1123}    & \textbf{4.65}  & 1156  & 7.74 \\
		A-n63-k10 & 62  & 1314  & \textbf{1396}  & \textbf{6.24}    & 1519    & 15.60 & 1416  & 7.76 \\
            A-n64-k9  & 63  & 1401  & 1517  & 8.28    & 1659    & 18.40 & \textbf{1493}  & \textbf{7.76} \\
            A-n69-k9  & 68  & 1159  & \textbf{1216}  & \textbf{4.92}    & 1264    & 9.05  & 1237  & 6.73 \\
            B-n34-k5  & 33  & 788   & 867   & 10.03    & \textbf{812}     & \textbf{3.04}  & 837   & 6.22 \\
            B-n35-k5  & 34  & 955   & \textbf{962}   & \textbf{0.73}    & 986     & 3.24  & 1005  & 5.24 \\
            B-n45-k6  & 43  & 678   & \textbf{706}   & \textbf{4.13}    & 729     & 7.52  & 755   & 11.36\\
            B-n51-k7  & 50  & 1032  & 1240  & 20.22   & \textbf{1045}    & \textbf{1.25}  & 1173  & 13.66\\
            P-n50-k8  & 49  & 631   & \textbf{650}   & \textbf{3.01}    & 655     & 3.80  & 660   & 4.60 \\
            P-n51-k10 & 50  & 741   & 806   & 8.77    & 811     & 9.44  & \textbf{773}   & \textbf{4.32} \\
            P-n70-k10 & 69  & 827   & 877   & 6.04    & \textbf{865}     & \textbf{4.59}  & 900   & 8.83 \\
            
            \cdashline{1-9}[0.8pt/2pt]
		{Average Gap}  & - & - & - & \textbf{5.74} & - & 6.09  & - & 7.21  \\
		\bottomrule
	\end{tabular}
 }
\end{table*}

\section{Conclusion}
In this paper, we introduce a framework for solving vehicle routing problems using graph attention sampling-based learning. Our framework automatically selects the top $K$\% correlated nodes as an encoder input to generate high-quality embeddings for nodes, edges, and the whole graph. We then use a multi-head attention based decoder that utilizes the embedding representation to construct the solution via a policy-driven deep reinforcement learning schema. We conducted extensive experiments on both randomly generated VRP instances and benchmark datasets and compared our results with six baseline methods to demonstrate the effectiveness of our proposed framework. Our analysis revealed that the proposed model achieved new state-of-the-art performance on CVRP tasks compared to other data-driven end-to-end methods. It is worth noting that purely data-driven reinforcement learning methodologies exhibit a high degree of sensitivity to the dependence on the underlying data distribution. An exaggerated imbalance in the distribution of nodes and capacity is unlikely to yield improved outcomes. Consequently, end-to-end learning strategies that can effectively account for various data distributions may emerge as a promising research direction in the future.

\section*{Acknowledgments}
This work was supported in part by the National Natural Science Foundation of China under Grant 72071116 and in part by the Ningbo Municipal Bureau of Science and Technology under Grant 2021Z173.

\bibliography{main}

\begin{thebibliography}{39}
\providecommand{\natexlab}[1]{#1}
\providecommand{\url}[1]{{#1}}
\providecommand{\urlprefix}{URL }
\providecommand{\doi}[1]{\url{https://doi.org/#1}}
\providecommand{\eprint}[2][]{\url{#2}}
 \bibcommenthead

\bibitem[{Bai et~al(2023)Bai, Chen, Chen, Cui, Gong, He, Jiang, Jin, Jin, Kendall et~al}]{bai2023analytics}
Bai R, Chen X, Chen ZL, et~al (2023) Analytics and machine learning in vehicle routing research. International Journal of Production Research 61(1):4--30

\bibitem[{Bello et~al(2016)Bello, Pham, Le, Norouzi, and Bengio}]{belloNeuralCombinatorialOptimization2017}
Bello I, Pham H, Le QV, et~al (2016) Neural combinatorial optimization with reinforcement learning. arXiv preprint arXiv:161109940

\bibitem[{Braekers et~al(2016)Braekers, Ramaekers, and Van~Nieuwenhuyse}]{braekers2016vehicle}
Braekers K, Ramaekers K, Van~Nieuwenhuyse I (2016) The vehicle routing problem: State of the art classification and review. Computers \& industrial engineering 99:300--313

\bibitem[{Chen et~al(2020)Chen, Qu, Bai, and Laesanklang}]{chen2020variable}
Chen B, Qu R, Bai R, et~al (2020) A variable neighborhood search algorithm with reinforcement learning for a real-life periodic vehicle routing problem with time windows and open routes. RAIRO - Operations Research 54(5):1467--1494

\bibitem[{Feng et~al(2021)Feng, Huang, Zhou, Zhong, Gupta, Tang, and Tan}]{fengExplicitEvolutionaryMultitasking2021b}
Feng L, Huang Y, Zhou L, et~al (2021) Explicit evolutionary multitasking for combinatorial optimization: A case study on capacitated vehicle routing problem. IEEE Transactions on Cybernetics 51(6):3143--3156

\bibitem[{Fey and Lenssen(2019)}]{feyFastGraphRepresentation2019}
Fey M, Lenssen JE (2019) Fast graph representation learning with pytorch geometric. arXiv preprint arXiv:190302428

\bibitem[{Glorot and Bengio(2010)}]{glorot2010understanding}
Glorot X, Bengio Y (2010) Understanding the difficulty of training deep feedforward neural networks. In: Proceedings of the thirteenth international conference on artificial intelligence and statistics, JMLR Workshop and Conference Proceedings, pp 249--256

\bibitem[{{Google Optimization Tools}(2024)}]{ortools}
{Google Optimization Tools} (2024) Google optimization tools. Online, \urlprefix\url{https://developers.google.com/optimization/}, accessed: 2024-03-26

\bibitem[{{Gurobi Optimization, LLC}(2024)}]{Gurobi2024}
{Gurobi Optimization, LLC} (2024) Gurobi optimizer. \urlprefix\url{https://www.gurobi.com/}, accessed: 2024-03-26

\bibitem[{Hamilton et~al(2017)Hamilton, Ying, and Leskovec}]{hamiltonInductiveRepresentationLearning2017a}
Hamilton W, Ying Z, Leskovec J (2017) Inductive representation learning on large graphs. In: Advances in Neural Information Processing Systems, vol~30. Curran Associates, Inc.

\bibitem[{Ioffe and Szegedy(2015)}]{ioffe2015batch}
Ioffe S, Szegedy C (2015) Batch normalization: Accelerating deep network training by reducing internal covariate shift. In: International conference on machine learning, pmlr, pp 448--456

\bibitem[{Joshi et~al(2020)Joshi, Cappart, Rousseau, and Laurent}]{joshiLearningTravellingSalesperson2022}
Joshi CK, Cappart Q, Rousseau LM, et~al (2020) Learning the travelling salesperson problem requires rethinking generalization. arXiv preprint arXiv:200607054

\bibitem[{Kalatzantonakis et~al(2023)Kalatzantonakis, Sifaleras, and Samaras}]{kalatzantonakisReinforcementLearningVariableNeighborhood2023a}
Kalatzantonakis P, Sifaleras A, Samaras N (2023) A reinforcement learning-variable neighborhood search method for the capacitated vehicle routing problem. Expert Systems with Applications 213:118812

\bibitem[{Khalil et~al(2017)Khalil, Dai, Zhang, Dilkina, and Song}]{khalilLearningCombinatorialOptimization2017}
Khalil E, Dai H, Zhang Y, et~al (2017) Learning combinatorial optimization algorithms over graphs. In: Advances in Neural Information Processing Systems, vol~30. Curran Associates, Inc.

\bibitem[{Kingma and Ba(2014)}]{kingma2014adam}
Kingma DP, Ba J (2014) Adam: A method for stochastic optimization. arXiv preprint arXiv:14126980

\bibitem[{Kipf and Welling(2016)}]{kipfSemiSupervisedClassificationGraph2017a}
Kipf TN, Welling M (2016) Semi-supervised classification with graph convolutional networks. arXiv preprint arXiv:160902907

\bibitem[{Kool et~al(2019)Kool, Van~Hoof, and Welling}]{koolAttentionLearnSolve2019}
Kool W, Van~Hoof H, Welling M (2019) Attention, learn to solve routing problems! arXiv preprint arXiv:180308475

\bibitem[{Kwon et~al(2020)Kwon, Choo, Kim, Yoon, Gwon, and Min}]{kwonPOMOPolicyOptimization2020}
Kwon YD, Choo J, Kim B, et~al (2020) Pomo: Policy optimization with multiple optima for reinforcement learning. In: Advances in Neural Information Processing Systems, vol~33. Curran Associates, Inc., pp 21188--21198

\bibitem[{Lauri et~al(2023)Lauri, Hsu, and Pajarinen}]{lauriPartiallyObservableMarkov2023}
Lauri M, Hsu D, Pajarinen J (2023) Partially observable markov decision processes in robotics: A survey. IEEE Transactions on Robotics 39(1):21--40

\bibitem[{Lei et~al(2022)Lei, Guo, Wang, Wu, and Zhao}]{leiSolveRoutingProblems2022}
Lei K, Guo P, Wang Y, et~al (2022) Solve routing problems with a residual edge-graph attention neural network. Neurocomputing 508:79--98

\bibitem[{Lin and Kernighan(1973)}]{linEffectiveHeuristicAlgorithm1973}
Lin S, Kernighan BW (1973) An effective heuristic algorithm for the traveling-salesman problem. Operations Research 21(2):498--516. {\href{https://arxiv.org/abs/169020}{{169020}}}

\bibitem[{Liu et~al(2023)Liu, Zhang, Tang, and Yao}]{liuHowGoodNeural2023a}
Liu S, Zhang Y, Tang K, et~al (2023) How good is neural combinatorial optimization? a systematic evaluation on the traveling salesman problem. IEEE Computational Intelligence Magazine 18(3):14--28

\bibitem[{Ma et~al(2019)Ma, Ge, He, Thaker, and Drori}]{maCombinatorialOptimizationGraph2019}
Ma Q, Ge S, He D, et~al (2019) Combinatorial optimization by graph pointer networks and hierarchical reinforcement learning. arXiv preprint arXiv:191104936

\bibitem[{Mnih et~al(2015)Mnih, Kavukcuoglu, Silver, Rusu, Veness, Bellemare, Graves, Riedmiller, Fidjeland, Ostrovski, Petersen, Beattie, Sadik, Antonoglou, King, Kumaran, Wierstra, Legg, and Hassabis}]{mnihHumanlevelControlDeep2015}
Mnih V, Kavukcuoglu K, Silver D, et~al (2015) Human-level control through deep reinforcement learning. Nature 518(7540):529--533

\bibitem[{Mnih et~al(2016)Mnih, Badia, Mirza, Graves, Lillicrap, Harley, Silver, and Kavukcuoglu}]{mnihAsynchronousMethodsDeep}
Mnih V, Badia AP, Mirza M, et~al (2016) Asynchronous methods for deep reinforcement learning. In: International conference on machine learning, PMLR, pp 1928--1937

\bibitem[{Nazari et~al(2018)Nazari, Oroojlooy, Snyder, and Tak{\'a}c}]{nazariReinforcementLearningSolving2018}
Nazari M, Oroojlooy A, Snyder L, et~al (2018) Reinforcement learning for solving the vehicle routing problem. Advances in neural information processing systems 31

\bibitem[{Ng et~al(2017)Ng, Lee, Zhang, Wu, and Ho}]{ngMultipleColoniesArtificial2017}
Ng K, Lee C, Zhang S, et~al (2017) A multiple colonies artificial bee colony algorithm for a capacitated vehicle routing problem and re-routing strategies under time-dependent traffic congestion. Computers \& Industrial Engineering 109:151--168

\bibitem[{Sutton et~al(1999)Sutton, McAllester, Singh, and Mansour}]{suttonPolicyGradientMethods1999}
Sutton RS, McAllester D, Singh S, et~al (1999) Policy gradient methods for reinforcement learning with function approximation. In: Advances in Neural Information Processing Systems, vol~12. MIT Press

\bibitem[{Szegedy et~al(2017)Szegedy, Ioffe, Vanhoucke, and Alemi}]{szegedyInceptionv4InceptionResNetImpact2017}
Szegedy C, Ioffe S, Vanhoucke V, et~al (2017) Inception-v4, inception-resnet and the impact of residual connections on learning. Proceedings of the AAAI Conference on Artificial Intelligence 31(1)

\bibitem[{Toth and Vigo(2014)}]{toth2014vehicle}
Toth P, Vigo D (2014) Vehicle routing: problems, methods, and applications. SIAM

\bibitem[{Uchoa et~al(2017)Uchoa, Pecin, Pessoa, Poggi, Vidal, and Subramanian}]{uchoa2017new}
Uchoa E, Pecin D, Pessoa A, et~al (2017) New benchmark instances for the capacitated vehicle routing problem. European Journal of Operational Research 257(3):845--858

\bibitem[{Vaswani et~al(2017)Vaswani, Shazeer, Parmar, Uszkoreit, Jones, Gomez, Kaiser, and Polosukhin}]{vaswaniAttentionAllYou2023}
Vaswani A, Shazeer N, Parmar N, et~al (2017) Attention is all you need. Advances in neural information processing systems 30

\bibitem[{Veli{\v{c}}kovi{\'c} et~al(2017)Veli{\v{c}}kovi{\'c}, Cucurull, Casanova, Romero, Lio, and Bengio}]{velickovicGraphAttentionNetworks2018a}
Veli{\v{c}}kovi{\'c} P, Cucurull G, Casanova A, et~al (2017) Graph attention networks. arXiv preprint arXiv:171010903

\bibitem[{Vinyals et~al(2015)Vinyals, Fortunato, and Jaitly}]{vinyalsPointerNetworks2017}
Vinyals O, Fortunato M, Jaitly N (2015) Pointer networks. Advances in neural information processing systems 28

\bibitem[{Xue et~al(2021)Xue, Bai, Qu, and Aickelin}]{xue2021hybrid}
Xue N, Bai R, Qu R, et~al (2021) A hybrid pricing and cutting approach for the multi-shift full truckload vehicle routing problem. European Journal of Operational Research 292(2):500--514. \doi{10.1016/j.ejor.2020.10.037}, publisher Copyright: {\textcopyright} 2020 Elsevier B.V.

\bibitem[{Yang et~al(2021)Yang, Ke, Wang, and Lam}]{yangBranchpriceandcutAlgorithmVehicle2021}
Yang W, Ke L, Wang DZ, et~al (2021) A branch-price-and-cut algorithm for the vehicle routing problem with release and due dates. Transportation Research Part E: Logistics and Transportation Review 145:102167

\bibitem[{Zaremba et~al(2014)Zaremba, Sutskever, and Vinyals}]{zarembaRecurrentNeuralNetwork2015}
Zaremba W, Sutskever I, Vinyals O (2014) Recurrent neural network regularization. arXiv preprint arXiv:14092329

\bibitem[{Zhang et~al(2021)Zhang, Liu, Chen, Huang, Zhu, and Zheng}]{zhangGPSPolicydrivenSampling2022}
Zhang T, Liu Y, Chen X, et~al (2021) Gps: A policy-driven sampling approach for graph representation learning. arXiv preprint arXiv:211214482

\bibitem[{Zhao et~al(2021)Zhao, Mao, Zhao, and Zou}]{zhaoHybridDeepReinforcement2021}
Zhao J, Mao M, Zhao X, et~al (2021) A hybrid of deep reinforcement learning and local search for the vehicle routing problems. IEEE Transactions on Intelligent Transportation Systems 22(11):7208--7218

\end{thebibliography}

\end{document}